\documentclass[11pt]{article}

\usepackage[final]{acl}

\usepackage{times}
\usepackage{latexsym}

\usepackage[T1]{fontenc}
\usepackage[utf8]{inputenc}

\usepackage{microtype}

\usepackage{inconsolata}

\usepackage{graphicx}
\usepackage{subcaption}
\usepackage{enumitem,amsmath}
\usepackage{multirow}
\usepackage{multicol}
\usepackage{algorithm}
\usepackage{algpseudocode}
\usepackage{amsmath}
\usepackage{amssymb}
\usepackage{booktabs}
\usepackage{array}
\usepackage{xltabular}
\usepackage{listings}
\newtheorem{definition}{Definition}
\title{ProcessLight: Process Supervision for Large Language Model Based Traffic Signal Control}

\author{
\textbf{Shuaitao Zhao},
\textbf{Tianlong Zhou},
\textbf{Weijie Wang},
\textbf{Jiasheng Shi},
\textbf{Weixiong Rao}\textsuperscript{}\thanks{Corresponding author.}
\\
School of Computer Science and Technology, Tongji University, Shanghai, China
\\
\texttt{
   {\{shuaitaozhao,wxrao\}@tongji.edu.cn}
 }
}

\begin{document}
\maketitle

\begin{abstract}
Large Language Models (LLMs) have recently been introduced into traffic signal control (TSC) as decision agents due to their strengths in human-readable reasoning generation.
Yet, existing LLM TSC methods optimize only from final outcomes and fail to distinguish valid from flawed reasoning steps, causing useful or misleading steps to be jointly updated and thus impairing the model’s learning of effective reasoning.
To bridge this gap, we propose an LLM-based framework ProcessLight to decompose signal decisions into verifiable semantic steps. Building on ProcessLight, we further develop Step-wise Traffic Process Policy Optimization (STeP-PO), a novel reinforcement learning framework that optimizes structured reasoning processes through step-level credit assignment.
Specifically, STeP-PO uses step quality scores to evaluate local reasoning quality and step importance to measure each step's influence on the final action, and then assigns step-level advantages over a semantic step tree structure. 
The resulting step-level advantages are propagated to reasoning tokens, enabling fine-grained policy optimization beyond outcome-only rewards. 
Extensive experiments over multiple real-world datasets demonstrate the superiority of our methods.
Our code is available at \href{https://github.com/wenzhaoabc/processlight}{https://github.com/wenzhaoabc/processlight}.
\end{abstract}

\section{Introduction}

\begin{figure}[t]
    \centering
    \includegraphics[width=1.0\linewidth]{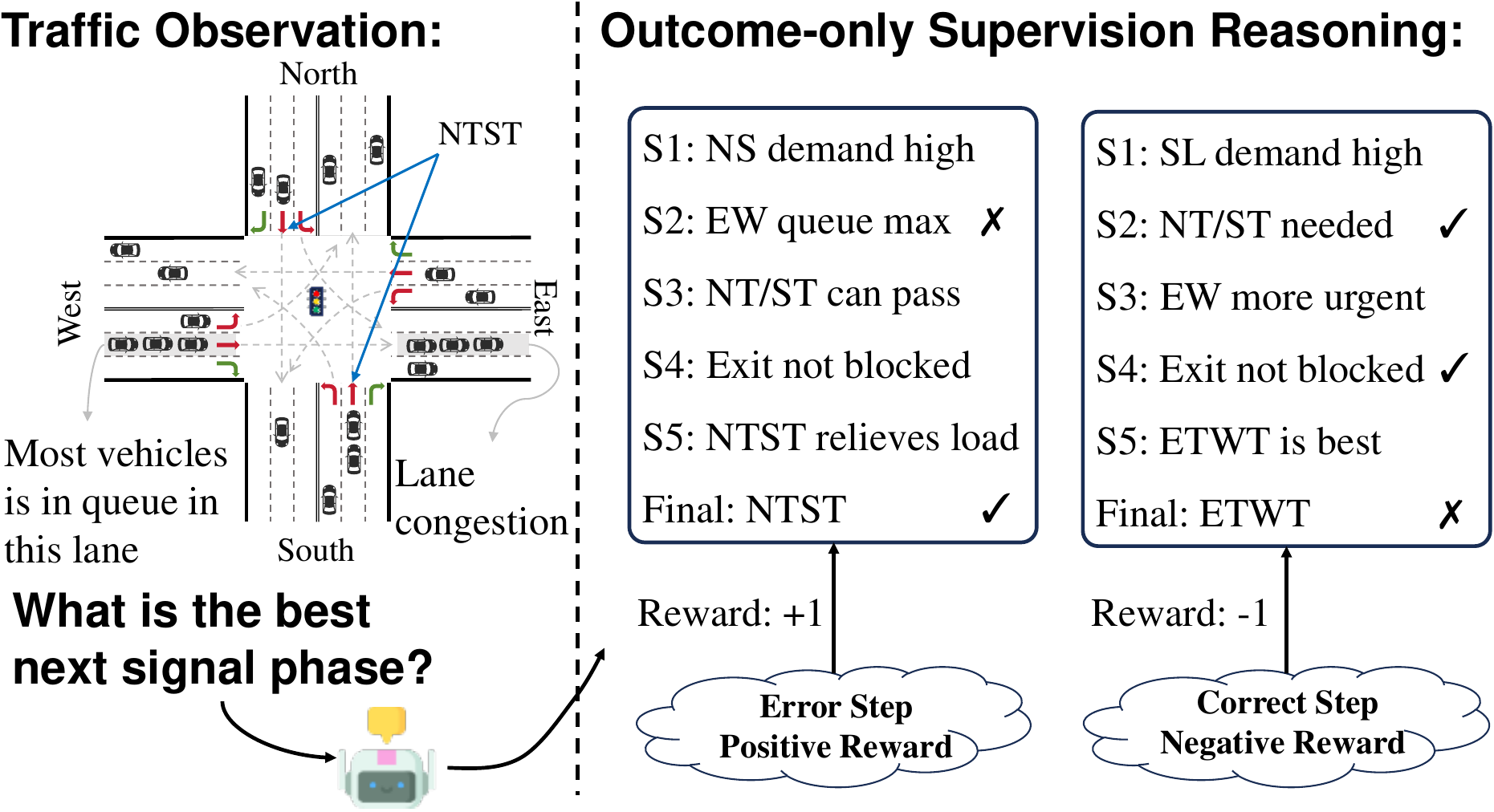}
    \caption{Outcome-level supervision cannot effectively distinguish correct and incorrect reasoning steps in the traffic signal reasoning process, weakening the model's ability to learn correct reasoning.}
    \label{fig:motivation}
\end{figure}

Rapid urbanization has led to a substantial increase in urban population, making traffic congestion an increasingly severe challenge for modern cities~\cite{survey,transformerlight,traffictimingmanual,xiaolei.mdm19}. Traffic signal control, as a core mechanism for regulating vehicle flows at intersections, has long been a central research problem in intelligent transportation systems.

Existing TSC methods have evolved from traditional controllers to deep reinforcement learning approaches. Rule-based controllers, such as fixed-time control and pressure-based control, are computationally efficient and highly interpretable. Yet, they heavily depend upon manually designed priors, and struggle to adapt to complex and dynamic traffic conditions~\cite{attendlight, presslight, liang2022oam, transformerlight}. Reinforcement learning(RL) based methods learn adaptive signal control policies from feedback provided by traffic simulators~\cite{intellilight,presslight,colight,attendlight,expression,transformerlight,bingquan.cikm19}. Despite promising performance, these methods are trained on compact numerical states and scalar rewards, which limits their interpretability and generalization ability in diverse traffic scenarios.

Recently, large language models (LLMs) have been introduced into traffic signal control to exploit their logical reasoning ability and natural-language generalization capability. Existing studies show that LLMs can transform traffic observations into interpretable control decisions~\cite{llmlight,collmlight,trafficr1}. However, current LLM-based TSC methods mainly optimize models according to final actions or overall trajectory outcomes, while largely overlooking the intermediate reasoning steps. As a result, they cannot effectively distinguish correct reasoning steps from erroneous ones. Erroneous intermediate reasoning may still be reinforced by a positive trajectory-level reward when the final action happens to be correct, thereby misleading the model to learn unreliable reasoning patterns.

As illustrated in Figure~\ref{fig:motivation}, even when a reasoning trajectory contains \emph{incorrect} intermediate steps, the entire trajectory may still be reinforced when the final control decision is correct. In contrast, when the final decision is incorrect, the whole reasoning process is penalized, even though some intermediate traffic judgments may be valid and useful. This outcome-only optimization paradigm provides coarse and potentially biased supervision, making it difficult to determine which reasoning steps should be reinforced, corrected, or suppressed.

To bridge this gap, we propose \textbf{ProcessLight}, an LLM-based traffic signal control framework with explicit step-wise reasoning. By decomposing each traffic signal decision into a sequence of reasoning steps, ProcessLight can provide verifiable traffic state evidence for each step, and thus improves the interpretability of LLM-based signal control. More importantly, it provides fine-grained process-level information for reasoning supervision.

We further propose \textbf{STeP-PO}, a Step-wise Traffic Process Policy Optimization algorithm for optimizing ProcessLight. STeP-PO consists of three key components. First, it evaluates the quality of each reasoning step based on traffic-grounded claims by combining semantic soft matching with observation-based validity verification. Second, it estimates step importance through counterfactual sensitivity, where each reasoning step is removed and the resulting change in the final action distribution is measured. Third, it constructs a semantic step tree that merges semantically equivalent reasoning steps under the same preceding logic and assigns comparable hierarchical advantages to different steps. These advantages are further propagated to the token level within the reasoning text, enabling fine-grained policy optimization beyond reliance on outcome-only rewards. In summary, we make the following contributions.

\begin{itemize}
    \item We propose \textbf{ProcessLight}, an LLM-based traffic signal control framework that formulates TSC as a process-supervised reasoning problem, enabling fine-grained optimization of intermediate traffic judgments.
    
    \item We introduce \textbf{STeP-PO}, a step-wise policy optimization algorithm that integrates step-quality evaluation and step-importance estimation to provide step-level supervision for LLM-based traffic signal control.
    
    \item We conduct extensive experiments in representative traffic simulation scenarios, demonstrating the effectiveness of the proposed method.
\end{itemize}

\section{Related Work}

\subsection{Traffic Signal Control}
Early methods of traffic signal control, e.g.,  fixed-time control and MaxPressure, are simple and stable, but their hand-crafted rules make them less adaptive to time-varying traffic demand and network-level coordination \cite{traffictimingmanual,maxpressure,qadrireview,survey}. Deep reinforcement learning (DRL) methods improve adaptivity by learning signal policies from simulation. Representative studies include PressLight, which integrates pressure-based control into RL \cite{presslight}, CoLight, which models network cooperation with graph attention \cite{colight}, and MPLight, which scales decentralized DRL to large road networks \cite{mplight}. However, these neural controllers remain limited by black-box decision processes, weak interpretability, and unstable generalization under distribution shifts. Recently, LLM-based controllers have introduced language reasoning into TSC. LLMLight treats LLMs as traffic signal agents \cite{llmlight}, CoLLMLight studies cooperative LLM agents for network-wide control \cite{collmlight}, and Traffic-R1 applies two-stage RL to reduce the gap between reasoning and deployment \cite{trafficr1}. These methods mainly optimize policies using final control outcomes or action-level feedback, leaving intermediate reasoning steps weakly supervised.

\subsection{RL Fine-tuning for LLMs}
RL fine-tuning is widely used to improve instruction following and reasoning. RLHF typically optimizes LLMs with PPO \cite{paul2017deep,ouyang2022training,ppo}, but the used critic for advantage estimation suffers from high memory cost and training instability. DPO bypasses explicit reward modeling by directly optimizing preference pairs \cite{DPO}, while GRPO removes the critic through group-relative baselines is effective for reasoning-oriented training \cite{deepseekmath,guo2025deepseek}. Nevertheless, trajectory-level rewards cannot distinguish the quality or contribution of individual reasoning steps within the same response. Recent credit-assignment methods, e.g., VinePPO and SPO, address this limitation with finer-grained advantage estimation \cite{vineppo,spo}. In contrast, ProcessLight introduces step-level reward design and credit assignment for TSC reasoning, enabling direct supervision of how an LLM derives traffic-signal actions rather than only whether the final action succeeds.

\section{Problem Definition}
\label{sec:preliminaries}

In this section, we introduce the key concepts of traffic signal control and formulate the LLM-based traffic signal control problem studied in this paper. More details on intersections, movements and signal phases are provided in the Appendix~\ref{app:concepts}.

\begin{definition}
\textbf{Road Network.}
The road network is a directed graph composed of intersections $I$ and roads $E$. Each road consists of multiple lanes $L$, and each lane corresponds to a movement at the intersection.
\end{definition}

\begin{definition}
\textbf{Traffic Signal Control}. At each signal-switching time step, the agent selects one phase $a$ from a predefined phase set $\mathcal{A} = \{a_1,\dotsc,a_m\}$. Each phase corresponds to a group of non-conflicting movements.
\end{definition}

We formulate traffic signal control as a partially observable Markov decision process. Each intersection is controlled by an LLM agent with policy $\pi_\theta$. At each signal-switching time step, the agent observes traffic conditions $\mathcal{O}$. Given the task description $\mathcal{D}_{\text{task}}$ and reasoning step guide $\mathcal{G}$, the LLM agent generates a step-by-step reasoning trajectory $Y$ and an executable phase action ${a}\in \mathcal{A}$ from the predefined phase set $\mathcal{A}$  by the policy $\pi_\theta$:
\begin{equation}
    (Y,{a})   % y reasoning path, a - action , y+a - trajectory
    =
    \pi_\theta
    \left(
    \text{Prompt}
    (\mathcal{O},\mathcal{D}_{\text{task}},\mathcal{G},\mathcal{A})
    \right).
\end{equation}

Here, the reasoning trajectory $Y$ contains $K$ steps $(s^1,\dots,s^K)$. The objective of our task is to optimize the policy $\pi_\theta$ to improve transportation efficiency through high-quality reasoning.

\section{Methodology}
\label{sec:method}

In this section, we first present an overview of ProcessLight, and then introduce the proposed Step-wise Traffic Process Policy Optimization (STeP-PO) that is used to optimize ProcessLight. 

\begin{figure}[t]
    \centering
    \includegraphics[width=1.0\linewidth]{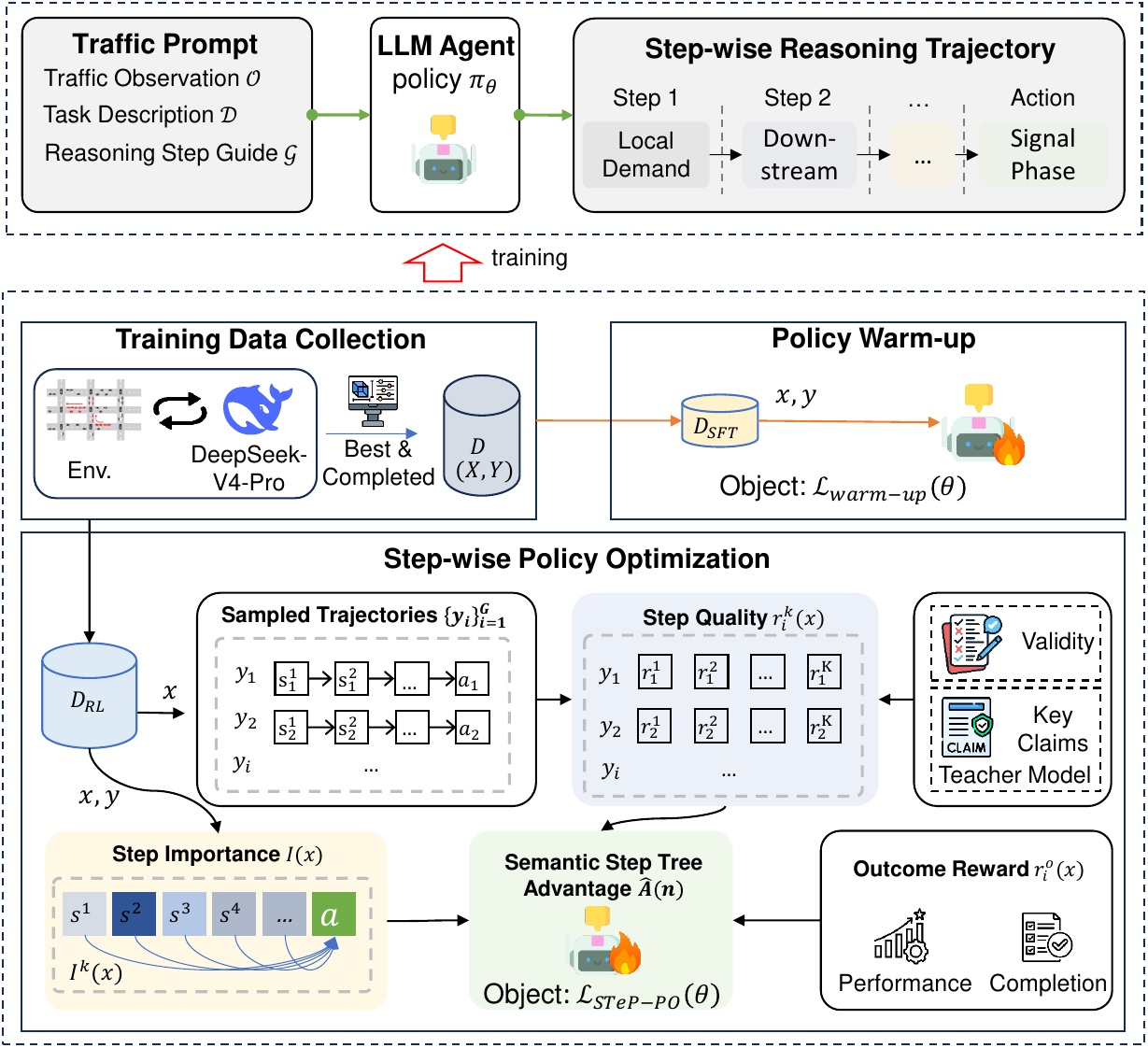}
    \caption{Overview of ProcessLight with two stages, (1) task prompt construction and (2) step-wise reasoning. The proposed STeP-PO algorithm is used for policy optimization.}
    \label{fig:overview}
\end{figure}

\subsection{ProcessLight Framework}
\label{subsec:processlight_framework}
In Figure~\ref{fig:overview}, the workflow of ProcessLight consists of two stages: (1) prompt construction and (2) step-wise reasoning and action execution.

\textit{Prompt Construction.} The prompt contains three main parts: current traffic observation $\mathcal{O}$, task description $\mathcal{D}_\text{task}$ and reasoning step guide $\mathcal{G}$. The observation $\mathcal{O}$ contains queue lengths and vehicle counts of the signalized and neighboring intersections, together with short-term historical trends for spatiotemporal reasoning. The guide $\mathcal{G}$ defines the reasoning direction and step decomposition, requiring the LLM to examine the current phases, local demand, downstream congestion, neighboring coordination, traffic trends, and phase fairness before selecting the final signal control action. We provide prompt examples in Appendix\ref{app:prompt}.

\textit{Step-wise Reasoning and Action Execution}. Given the  prompt $X$, ProcessLight performs step-wise reasoning by following the number $K$ of predefined decision steps. 
At each step {$s^k$ with $k\in [1, K]$,} the LLM agent generates both the reasoning evidence and the corresponding step-level conclusion, and then gives the final signal control action $a$. Hence, ProcessLight not only executes effective traffic control decisions $a$, but also provides fine-grained reasoning evidence for each decision {step $s^k$}. 
This step-wise reasoning process improves interpretability and, more importantly, provides the necessary step-level information for subsequent reasoning process supervision.

\subsection{STeP-PO}

\subsubsection{Overview}

This section introduces the proposed Step-wise Traffic Process Policy Optimization (STeP-PO). STeP-PO contains three stages: (1)\textit{Training Data Collection}, (2)\textit{Policy Warm-up} and (3)\textit{Step-wise Policy Optimization}. We provide the overall algorithm in Appendix~\ref{app:alg}.

\subsubsection{Training Data Collection}

We begin by collecting diverse traffic scenarios from the traffic simulator. 
%\textcolor{yellow}{For each scenario, we evaluate all phase actions through long-term network-wide simulation and select the phase action that yields the best traffic performance (i.e. lowest average queue length).} 
For each scenario, we evaluate all candidate phase actions for the target intersection with a fixed-horizon network-wide simulation, while other intersections follow the same baseline control policy, and select the action that yields the best traffic performance, e.g., lowest average queue length. Then, we prompt {a large language model }(e.g. DeepSeek-V4-Pro in Figure~\ref{fig:overview}) using the prompt construction process described in Section~\ref{subsec:processlight_framework} to generate step-wise reasoning trajectories aligned with the best action and the {required output format}. 
The resulting dataset $\mathcal{D}$ consists of input prompts $X$ paired with step-wise reasoning trajectories $Y$. A small subset $\mathcal{D}_{\mathrm{SFT}}\subseteq \mathcal{D}$ is used for supervised warm-up, while the remaining data $\mathcal{D}_{\mathrm{RL}}$ for step-wise policy optimization.

\subsubsection{Policy Warm-up}

To equip the policy model with traffic-aware reasoning and format-following capabilities, we introduce a supervised warm-up stage before reinforcement learning (RL). The warm-up fine-tuning is performed on the above subset $\mathcal{D}_{\mathrm{SFT}}$ with the following objective:

\small
\begin{equation}\small
    \mathcal{L}_{\mathrm{warm-up}}(\theta)
    =-\mathbb{E}_{(X,Y)\sim \mathcal{D}_{\mathrm{SFT}}}
    \left[
    \sum_{t=1}^{|Y|}
    \log \pi_{\theta}\left(Y_t \mid X, Y_{<t}\right)
    \right]
    \label{eq:warmup_loss}
\end{equation}
\normalsize

\subsubsection{Step-wise policy optimization}

Outcome-only optimization is performed on entire reasoning trajectories with trivial guidance on intermediate steps. STeP-PO addresses this issue by optimizing the policy $\pi_{\theta}$ on $\mathcal{D}_{\mathrm{RL}}$ with a step-wise objective $\mathcal{L}_{\mathrm{STeP\text{-}PO}}(\theta)$. As shown in Fig.~\ref{fig:overview}, STeP-PO involves three components: \emph{step quality}, which measures whether an intermediate process is locally beneficial, \emph{step importance}, estimating the step contribution to the final action, and a \emph{semantic step tree}, where shared prefix steps are merged across trajectories into a single tree node. Hence, each node is assigned with a step-level advantage with a low variance. Step-level advantage is then assigned to the tokens within each step to derive token-level advantages for policy optimization.

\textbf{Step Quality.}
\label{subsubsec:reward_design}
Step quality helps to provide fine-grained supervision for intermediate reasoning steps, enabling STeP-PO to distinguish reliable local decisions from noisy or hallucinated ones. 
Specifically, for each prompt $X \in \mathcal{D}_{\mathrm{RL}}$, the policy model $\pi_{\theta}$ samples $G$ reasoning trajectories $Y=\{y_i\}_{i=1}^{G}$, where each trajectory $y_i$ is decomposed into $K$ steps, i.e., $y_i=(s_i^1,\ldots,s_i^K)$. 
To evaluate the decision quality of each step, we require the LLM to output step-level decision claims. 
Each claim is a fine-grained and verifiable traffic judgment grounded in the current observation, such as ``[ETWT][Queue Length] $>$ [NTST][Queue Length]'' or ``MAX[AWT][PHASE]=NLSL''. 
For the $k$-th step each training sample $y_i$, {a teacher model} pre-extracts the reference key-claim set, {denoted by $\mathcal{C}_i^k$}, to serve as the criterion for step-quality evaluation.

To estimate the step quality score, we use a soft key-claim matching method inspired by work~\cite{r1vl} and a traffic observation-based verification. Specifically, for the $k$-th step of trajectory $y_i$ and the associated claim set $C_i^k$, we first use a validity gate $\mathrm{Valid}(\cdot)$ to check whether all generated claims are supported by the current traffic observation.  {If any claim is contradicted by the observation}, the step quality score is assigned -1; otherwise, its reward is given by the matching score against the reference key claims:
\begin{equation}\small
\label{eq:step_reward}
r_i^k =
\begin{cases}
-1, & \text{if } \mathrm{Valid}(C_i^k)=\mathrm{False},\\
m_i^k, & \text{otherwise},
\end{cases}
\end{equation}
where {$m_i^k=|C_{i,\mathrm{match}}^k|/|\mathcal{C}_i^k|$ and $C_{i,\mathrm{match}}^k$ denotes the subset of key-claims matched by the generated claims}. 
This design penalizes hallucinated traffic judgments, supports semantically equivalent claim forms, and prevents reward inflation from irrelevant claims. It thereby provides a reliable step quality signal for step-wise policy optimization.

\textbf{Step Importance.}
\label{subsec:importance}
To assess the marginal effect of each reasoning step on the final decision, we incorporate counterfactual sensitivity into STeP-PO for step importance estimation. For each training pair $(x,y)\in\mathcal{D}_{\mathrm{RL}}$ with $y=(s^1,\dots,s^K)$, we remove the $k$-th step and construct the counterfactual trajectory $y^{-k}=(s^1,\dots,s^{k-1},s^{k+1},\dots,s^K)$. 
We further map each phase action to a dedicated action token (e.g., \texttt{ETWT} is presented as {a signal token}), such that phase probability can be directly obtained from next-token prediction probabilities.

We denote the phase distribution conditioned on the complete trajectory $y$ as $p^0(\cdot)$ and the counterfactual distribution conditioned on $y^{-k}$ as $p^{-k}(\cdot)$. 
The distributional discrepancy between $p^0(\cdot)$ and $p^{-k}(\cdot)$ measures the sensitivity of the final phase decision to step $s^k$. 
We employ $D_{\mathrm{JSD}}$ (Jensen--Shannon divergence) to quantify the divergence between distributions:
\begin{equation}\small
\label{eq:step_sensitivity}
b^k(x) =
D_{\mathrm{JSD}}\left(p^0(\cdot), p^{-k}(\cdot)\right).
\end{equation}
Then we perform normalization over trajectory steps to derive the relative importance:
\begin{equation}\small
\label{eq:step_imp}
I^k(x) =
\frac{b^k(x)}
{\sum_{u=1}^{K} b^u(x) + \epsilon},
\end{equation}
where the small constant $\epsilon$ prevents numerical instability. This normalization captures the relative effect of each reasoning step within the trajectory $y$. Thus, it  allows outcome feedback to be concentrated on those steps that are more decisive for the final phase action. Practically, these importance scores are pre-computed per prompt before policy optimization, and then reused across all associated rollouts. Thus, no additional forward or backward passes are required during RL training.

\noindent\textbf{Reward Design.}
We define the outcome and process rewards. The outcome reward $r_i^o$ consists of accuracy reward $r_i^{\mathrm{acc}}$ and reasoning completion reward $r_i^{\mathrm{com}}$. 
The former reward $r_i^{\mathrm{acc}}$ evaluates whether the model selects the best phase action, while the latter one $r_i^{\mathrm{com}}$ checks whether the model produces all required reasoning steps in the prescribed format. We denote $r_i^o=r_i^{\mathrm{acc}} + r_i^{\mathrm{com}}$. For process-level supervision, we directly use the step quality score of each step as the step reward.

\begin{figure}
    \centering
    \includegraphics[width=1.0\linewidth]{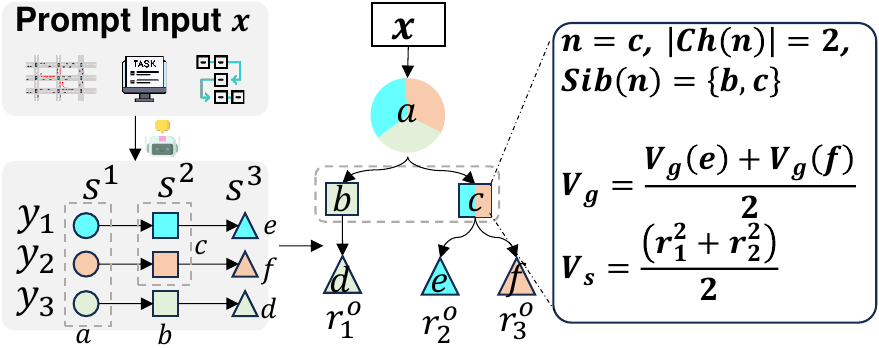}
    \caption{Semantic step tree construction. Reasoning steps with soft-matched claims are merged into the same node. The global state value is recursively propagated from child nodes, while the local step-level state value is computed from the step quality rewards.}
    \label{fig:tree}
\end{figure}

\textbf{Semantic Step Tree Advantage.}
\label{subsubsec:semantic_step_tree}
After evaluating step quality and deriving step importance with STeP-PO, we assign step-level advantages to strengthen supervision for intermediate reasoning. 
To this end, we adopt a tree-based structure to preserve prefix-conditioned dependencies during LLM reasoning and then to establish a fair local baseline for cross-step comparison.
% In Figure \ref{fig:tree}, we take the input $x$ as the root node, and \textcolor{purple}{then merge those steps across trajectories that belong to the same reasoning position and contain claims supported by identical soft matches, with final actions serving as leaf nodes}. 
In Figure \ref{fig:tree}, we take the input x as the root node, and then merge those reasoning steps that are located at the same reasoning position across different trajectories and have semantically‑matched claim sets.
Next, we attach each trajectory’s final action as a leaf node. Detailed algorithm for how to construct the semantic step tree is provided in Appendix~\ref{app:tree_construction}.

For a certain node $n$ within this tree structure, we denote its sibling nodes and child nodes by $\mathrm{Sib}(n)$ and $\mathrm{Chd}(n)$, respectively, and the set of reasoning steps contained in node $n$ by $\mathcal{S}(n)$. Next, we define the state value of node $n$ as $V(n)$, which consists of two parts: the global state value $V_{\mathrm{g}}(n)$, which indicates its contribution to the final outcome, and the step-level state value $V_{\mathrm{s}}(n)$, which measures the intrinsic quality of the corresponding reasoning step. Specifically, we recursively compute the global state value by bottom-up aggregation:
\begin{equation}\small
\hat{V}_{g}(n)=
\begin{cases}
r^{\mathrm{o}}_i, &\text{leaf node},\\
\frac{1}{|\mathrm{Chd}(n)|}\sum_{n' \in \mathrm{Chd}(n)} \hat{V}_{g}(n'), & \text{otherwise}.
\end{cases}
\end{equation}

For the step-level state value, we aggregate the step quality rewards of all reasoning steps merged into the same node:
\begin{equation}\small
\hat{V}_{\mathrm{s}}(n)
=
\frac{1}{|\mathcal{S}(n)|}
\sum_{s_i^k \in \mathcal{S}(n)} r_i^k .
\end{equation}

We then estimate the node-level advantage by jointly considering the global advantage and the step-level advantage. The global and step advantage are both computed with a normalized local baseline over sibling nodes, for $\ell \in \{\mathrm{g}, \mathrm{s}\}$:
\begin{equation}\small\small
\hat{A}_{\mathrm{\ell}}(n)
=
\frac{
\hat{V}_{\ell}(n)-\mathrm{mean}_{n' \in \mathrm{Sib}(n)} \hat{V}_{\ell}(n')
}{
\mathrm{std}_{n' \in \mathrm{Sib}(n)} \hat{V}_{\ell}(n')+\epsilon
},
\end{equation}
where $\epsilon$ is a small constant for numerical stability. Then we get global advantage $\hat{A}_{g}(n)$ and local advantage $\hat{A}_{s}(n)$. The final node-level advantage is then defined as
\begin{equation}\small
\label{eq:step_adv}
\hat{A}(n) = w^k(x)\hat{A}_{\mathrm{g}}(n) + \lambda \hat{A}_{\mathrm{s}}(n),
\end{equation}
where $\lambda$ is a hyperparameter controlling the contribution of the step-level advantage. The weight $w^k(x)$ is derived from the normalized step importance $I^k(x)$ to preserve the average scale of the global credit, where $\eta$ controls the strength of importance weighting.
\begin{equation}\small
\label{eq:weight}
w^k(x) = (1-\eta)+\eta K I^k(x).
\end{equation}

Unlike the conventional outcome-level advantage, the tree-structured advantage constructs comparable sibling node groups under shared reasoning prefixes and jointly considers local relative advantages according to step-level decision quality. Thus, STeP-PO alleviates the credit assignment bias caused by relying solely on overall reasoning outcomes, and  instead provide step-level optimization signals that explicitly preserve the contextual dependencies of preceding reasoning steps.

\subsubsection{Policy Optimization Objective}

We assign $\hat{A}(n)$ to all tokens within the corresponding reasoning-step segment and denote $\hat{A}_{i,t}$ as the advantage assigned to the $t$-th token in the $i$-th sampled trajectory $y_i$. Given a group of $G$ rollouts sampled from the old policy $\pi_{\theta_{\mathrm{old}}}$, we define the token-level importance sampling ratio and its clipped form as
\begin{equation}\small
\begin{aligned}
\rho_{i,t}(\theta)
&=
\frac{
\pi_{\theta}(y_{i,t} \mid x, y_{i,<t})
}{
\pi_{\theta_{\mathrm{old}}}(y_{i,t} \mid x, y_{i,<t})
},\\
\bar{\rho}_{i,t}(\theta)
&=\operatorname{clip}
(\rho_{i,t}(\theta), 1-\epsilon, 1+\epsilon).
\end{aligned}
\end{equation}

The final STeP-PO objective is defined as
\begin{equation}\small
\begin{aligned}
&\mathcal{L}_{\mathrm{STeP\text{-}PO}}(\theta)
= -
\mathbb{E}_{x,\{y_i\}_{i=1}^{G}}
\Bigg[
\frac{1}{G}
\sum_{i=1}^{G}
\frac{1}{|y_i|}
\sum_{t=1}^{|y_i|} \\
&\Big(
\min \big(
\rho_{i,t}(\theta)\hat A_{i,t},
\bar{\rho}_{i,t}(\theta)\hat A_{i,t}
\big)
- \beta D_{\mathrm{KL}}^{i,t}
\Big)
\Bigg],
\end{aligned}
\label{eq:step_po_obj}
\end{equation}
where $\beta$ controls the KL regularization strength. Following GRPO-style optimization, the token-level KL penalty is estimated by
\begin{equation}\small
D_{\mathrm{KL}}^{i,t}
=
\frac{
\pi_{\mathrm{ref}}(y_{i,t} \mid x,y_{i, <t})
}{
\pi_{\theta}(y_{i,t} \mid x,y_{i,<t})
}
-
\log
\frac{
\pi_{\mathrm{ref}}(y_{i,t} \mid x,y_{i,<t})
}{
\pi_{\theta}(y_{i,t} \mid x,y_{i,<t})
}
-1 .
\end{equation}

\section{Experiments}
\label{sec:experiments}

\subsection{Experimental Setup}

\subsubsection{Datasets}
%\textbf{Datasets}. 
As shown in Table~\ref{tab:dataStats}, we train our model on one synthetic dataset ({Synthetic-1}) and evaluate its zero-shot generalization ability on five real datasets (Jinan1, Jinan2, Jinan3, Hangzhou1, and Hangzhou2)~\cite{survey}. Detailed information about these datasets is provided in Appendix~\ref{app:datasets}.

% \begin{itemize}
%     \item \textbf{Jinan}: a 12-intersection structure from the Dongfeng sub-district in Jinan, China with three traffic flow traces. Every intersection is composed of two 400-meter east-west roads and two 800-meter north-south roads.
    
%     \item \textbf{Hangzhou}: a 16-intersection structure from the Gudang sub-district in Hangzhou, China, with two traffic flow datasets. Every intersection is composed of two 800-meter east-west roads and two 600-meter north-south roads.

%     \item \textbf{Synthetic-1}: A synthetic dataset consisting of 16 intersections and each road between two intersections is 300 meters long.
% \end{itemize}

\begin{table}[t]
\renewcommand{\arraystretch}{1.1}
 \small
 \centering
 \caption{Statistics of datasets.}
 \resizebox{0.48\textwidth}{!}{
  \begin{tabular}{c|c|c|cccc}
   \toprule
   \multirow{2}{*}{Flow dataset} & \multirow{2}{*}{Structure} & \multirow{2}{*}{Vehicles} & \multicolumn{4}{c}{Arrival rate (vehicles/5min)} \\
   \cline{4-7}
   &  & & Mean & Std & Max & Min \\
   \midrule
   Jinan 1 & \multirow{3}{*}{3$\times$4} & 6295 & 523.67 & 98.52 & 671 & 255 \\
   Jinan 2 & & 4365 & 362.83 & 74.81 & 493 & 236 \\
   Jinan 3 & & 5494 & 456.92 & 46.38 & 544 & 362 \\
   \midrule
   Hangzhou 1 & \multirow{2}{*}{4$\times$4} & 2983 & 247.67 & 40.44 & 332 & 211\\
   Hangzhou 2 & & 6984 & 581.08 & 318.43 & 1145 & 202 \\
   \midrule
   Synthetic 1 & {4$\times$4} & 8000 & 666.6 & 32.10 & 735 & 612 \\
   \bottomrule
  \end{tabular}}
  \label{tab:dataStats}

\end{table}

\begin{table*}[t]
\centering
\caption{Zero-shot traffic control performance on real-world traffic flows. Lower values indicate better performance. The best results are in bold and the second-best results are \underline{underlined}.}% All methods are evaluated under the same default random seed.}
\label{tab:main_results}
\resizebox{\textwidth}{!}{\Large
\begin{tabular}{lccccccccccccccc}
\toprule
\multirow{2}{*}{Method} 
& \multicolumn{3}{c|}{Jinan 1}
& \multicolumn{3}{c|}{Jinan 2}
& \multicolumn{3}{c|}{Jinan 3}
& \multicolumn{3}{c|}{Hangzhou 1}
& \multicolumn{3}{c}{Hangzhou 2} \\
\cmidrule(lr){2-4}
\cmidrule(lr){5-7}
\cmidrule(lr){8-10}
\cmidrule(lr){11-13}
\cmidrule(lr){14-16}
& ATT$\downarrow$ & AWT$\downarrow$ & AQL$\downarrow$
& ATT$\downarrow$ & AWT$\downarrow$ & AQL$\downarrow$
& ATT$\downarrow$ & AWT$\downarrow$ & AQL$\downarrow$
& ATT$\downarrow$ & AWT$\downarrow$ & AQL$\downarrow$
& ATT$\downarrow$ & AWT$\downarrow$ & AQL$\downarrow$ \\
\midrule
\multicolumn{16}{c}{\textbf{Traditional Methods}} \\
\midrule
FixedTime & 410.07 & 157.00 & 190.36 & 521.91 & 232.12 & 405.89 & 463.07 & 184.99 & 282.31 & 564.91 & 204.65 & 169.57 & 553.78 & 137.69 & 267.12 \\
MaxPressure & 305.15 & 59.93 & 72.67 & 318.12 & 77.98 & 136.36 & 298.02 & 62.23 & 94.96 & 343.97 & 47.08 & 39.01 & 438.54 & 102.06 & 197.99 \\
\midrule
\multicolumn{16}{c}{\textbf{RL Methods}} \\
\midrule
EfficientPressLight & 316.85 & 77.93 & 136.27 & 310.08 & 66.08 & 80.12 & 302.16 & 67.20 & 102.56 & 472.23 & 178.99 & 148.31 & 502.59 & 172.35 & 334.37 \\
EfficientCoLight & 310.23 & 70.10 & 122.57 & 298.99 & 54.05 & \underline{65.54} & 293.58 & 57.86 & 88.30 & 340.52 & 43.85 & 36.34 & 419.34 & 82.13 & 159.34 \\
AttendLight & 557.35 & 292.16 & 510.88 & 443.58 & 193.23 & 234.29 & 531.52 & 269.35 & 411.06 & 486.69 & 182.19 & 150.96 & 560.49 & 161.61 & 313.53 \\
EfficientMPLight & 311.41 & 71.43 & 124.91 & 301.97 & 57.28 & 69.45 & 293.78 & 58.04 & 88.58 & 384.45 & 93.46 & 77.44 & 433.5 & 92.14 & 178.76 \\
AdvancedCoLight & 298.81 & 60.40 & 105.61 & \textbf{292.99} & \textbf{49.01} & \textbf{59.41} & 286.18 & \underline{51.26} & \underline{78.23} & 328.58 & \underline{32.95} & \textbf{27.30} & \underline{407.93} & 76.50 & \underline{148.42} \\
MPLight & 466.00 & 214.83 & 375.65 & 384.79 & 138.23 & 167.61 & 421.48 & 185.16 & 282.57 & 558.10 & 234.91 & 194.65 & 550.89 & 151.67 & 294.24 \\
AdvancedMPLight & 313.15 & 75.18 & 131.47 & 298.2 & \underline{54.13} & 65.64 & 290.16 & 55.36 & 84.49 & 372.22 & 80.23 & 66.48 & 437.29 & 102.24 & 198.34 \\
CoLight & 478.90 & 224.42 & 392.42 & 407.75 & 158.29 & 191.93 & 429.4 & 188.05 & 286.98 & 522.52 & 209.19 & 173.34 & 535.23 & 145.58 & 282.42 \\
\midrule
\multicolumn{16}{c}{\textbf{LLM-based Methods}} \\
\midrule
LightGPT-3B & 406.13 & 160.26 & 194.32 & 411.45 & 173.48 & 303.34 & 402.70 & 167.71 & 255.95 & 412.47 & 117.12 & 97.05 & 503.97 & 137.13 & 266.03 \\
% CoLLMLight(Qwen2.5-3B) & 748.35 & 535.64 & 649.47 & 745.47 & 529.36 & 925.65 & 773.27 & 564.05 & 860.8 & 556.41 & 273.03 & 226.24 & 603.94 & 232.49 & 451.03  \\
% Traffic-R1(Qwen2.5-3B) & 356.72 & 113.06 & 137.08 & 382.88 & 143.98 & 251.76 & 352.67 & 118.1 & 180.23 & 390.05 & 96.42 & 79.89 & 476.89 & 125.26 & 243.01 \\ 

LightGPT-8B & 343.65 & 100.06 & 121.32 & 358.39 & 120.48 & 210.67 & 340.19 & 105.97 & 161.72 & 376.63 & 85.02 & 70.45 & 440.57 & 111.78 & 216.86 \\
LLMLight & 297.61 & 52.53 & 63.69 & 310.50 & 70.21 & 122.77 & 292.00 & 55.88 & 85.28 & 336.63 & 40.71 & 33.73 & 421.34 & 86.08 & 166.99 \\
CoLLMLight & 298.99 & 54.35 & 65.90 & 311.23 & 71.06 & 124.25 & 292.25 & 56.58 & 86.35 & 336.25 & 41.30 & 34.22 & 419.96 & \underline{74.02} & \textbf{143.59}  \\
Traffic-R1 & 299.75 & 54.44 & 66.01 & 313.04 & 72.73 & 127.19 & 293.71 & 57.65 & 87.98 & 334.89 & 38.44 & 31.85 & 421.31 & 80.66 & 156.49 \\

\midrule
\multicolumn{16}{c}{\textbf{ProcessLight (ours)}} \\
\midrule
Qwen2.5-3B        & 327.90 & 82.95 & 100.58 & 356.37 & 117.97 & 206.28 & 327.34 & 92.23 & 140.76 & 353.87 & 59.15 & 49.01 & 438.37 & 109.85 & 213.11 \\
Qwen3-8B          & 306.93 & 71.67 & 84.08  & 313.57 & 72.76  & 127.23 & 292.57 & 56.62 & 86.40  & 338.84 & 41.51 & 34.40 & 426.83 & 86.76 & 168.32 \\
Llama3.1-8B      & 317.47 & 80.05 & 94.84  & 339.83 & 115.69 & 189.50 & 327.03 & 89.59 & 128.86 & 324.80 & 56.14 & 42.87 & 413.31 & 97.75 & 202.24 \\
Qwen3-32B         & 299.67 & 54.37 & 65.92 & 312.16 & 71.67 & 125.33 & 293.96 & 57.85 & 88.28 & 338.05 & 40.87 & 33.86 & 426.62 & 85.51 & 168.88 \\
DeepSeek-V4-Flash & 296.83 & \underline{50.22} & 61.96 & 302.12 & 68.51 & 116.28 & 286.17 & 55.31 & 83.09 & \underline{321.98} & 41.64 & 34.51 & 416.45 & 82.60 & 164.13 \\
DeepSeek-V4-Pro  & \underline{290.93} & 51.33 & \underline{60.88} & 301.53 & 70.53 & \underline{114.83} & \underline{285.12} & 56.60 & 82.43 & 322.37 & 41.01 & 33.98 & 412.11 & 81.91 & 165.58 \\
\textbf{ProcessLight}
                  & \textbf{289.34} & \textbf{45.03} & \textbf{55.94} & \underline{293.18} & 54.16 & 102.67 & \textbf{284.33} & \textbf{50.97} & \textbf{78.20} & \textbf{314.99} & \textbf{32.82} & \underline{30.48} & \textbf{407.24} & \textbf{73.13} & 154.21 \\
\bottomrule
\end{tabular}
}
\end{table*}

\subsubsection{Environment Settings}
%\textbf{Environment Settings}. 
We conduct experiments on the open-source microscopic traffic simulator CityFlow~\cite{cityflow}. Each intersection adopts four signal phases, and right-turn traffic is always allowed. The green phase lasts 15 seconds, followed by a 3-second yellow light and a 2-second all-red interval. All traffic flow scenarios are simulated for one hour.

\subsubsection{Compared Methods}
%\textbf{Compared Methods}.
We adopt three following categories of evaluation methods. \textbf{Traditional transportation methods} include FixedTime~\cite{traffictimingmanual} and MaxPressure~\cite{maxpressure} as baselines. For \textbf{RL-based methods}, we compare with eight methods, including MPLight~\cite{mplight}, CoLight~\cite{colight}, AttendLight~\cite{attendlight}, EfficientPressLight~\cite{efficientpressure}, EfficientMPLight~\cite{efficientpressure}, EfficientCoLight~\cite{efficientpressure}, AdvancedMPLight~\cite{expression} and AdvancedCoLight~\cite{expression}. For \textbf{LLM-based methods}, we evaluate LLMLight~\cite{llmlight}, CoLLMLight~\cite{collmlight} and Traffic-R1~\cite{trafficr1}. Additionally, we assess the performance
of general LLMs integrated within our ProcessLight framework, including Llama, Qwen, and DeepSeek series. More details refer to Appendix~\ref{app:baselines}.

\subsubsection{Evaluation Metrics}
%\textbf{Evaluation Metrics}. 
We adopt three widely used traffic signal control metrics including \textit{Average Travel Time (ATT)}, \textit{Average Waiting Time (AWT)}, and \textit{Average Queue Length (AQL)}. Their lower values indicate better traffic efficiency and control performance. More details about these metrics refer to Appendix~\ref{app:metrics}.

\subsubsection{Implementation Details}
%\textbf{Implementation Details} 
We implement ProcessLight with Qwen2.5-3B~\cite{qwen25} as the LLM-agent and DeepSeek-V4-Pro~\cite{deepseekv4} as the teacher model, and split the collected dataset into two subsets: $\mathcal{D}_{\mathrm{SFT}}$ with 500 samples and 
$\mathcal{D}_{\mathrm{RL}}$ with 3000 samples. 
For fairness, all LLM baselines are trained and evaluated with the same experimental environment and configurations. With no availability of official Traffic-R1 source code, we have to adopt the model weights provided by Traffic-R1 HuggingFace \cite{trafficR1.url} for evaluation.

\subsection{Performance Comparison}
\label{sec:overall_performance}

We evaluate ProcessLight on conventional traffic signal control tasks using five real-world traffic flows from two China cities (Jinan and Hangzhou). Following the cross-domain evaluation protocol, all learning-based methods are trained on Synthetic-1 and then directly tested on unseen real-world traffic flows without further fine-tuning. As shown in Table~\ref{tab:main_results}, ProcessLight achieves the best overall performance among all compared methods, obtaining the best results on most metrics. It also consistently outperforms existing LLM-based traffic control methods on almost all metrics, demonstrating that optimizing the reasoning process provides a more effective control policy than relying only on prompted reasoning or outcome-level optimization.

{Without any fine-tuning}, the smaller Qwen2.5-3B surpasses LightGPT-8B across five datasets. This demonstrates that model performance is not exclusively determined by parameter scale and fine-tuning; the stepwise reasoning mechanism of ProcessLight also contributes. 
By compelling LLMs to generate explicit reasoning steps, ProcessLight enables LLM agents to comprehensively analyze core traffic control factors and make rational intermediate decisions, effectively boosting traffic control performance prior to policy optimization.

\subsection{Ablation Study}
\label{sec:ablation}

\begin{table}[t]
\centering
\caption{Ablation study on five real-world datasets.}
\label{tab:ablation}
\resizebox{\linewidth}{!}{\Huge
\begin{tabular}{lccccc}
\toprule
\multirow{2}{*}{Variant}
& \multicolumn{5}{c}{Average Travel Time (s) $\downarrow$} \\
\cmidrule(lr){2-6}
& Jinan 1 & Jinan 2 & Jinan 3 & Hangzhou 1 & Hangzhou 2 \\
\midrule
Qwen2.5-3B & 327.90 & 356.37 & 327.34 & 353.87 & 438.37 \\
+ Warm-up & 320.91 & 344.55 & 319.17 & 345.61 & 432.92 \\
+ SQ & 300.64 & 315.18 & 298.33 & 330.99 & 418.24 \\
+ SI & 293.23 & 312.87 & 292.18 & 318.72 & 409.82 \\
+ SSTA (ProcessLight) & \textbf{289.34} & \textbf{293.18} & \textbf{284.33} & \textbf{314.99} & \textbf{407.24} \\
\bottomrule
\end{tabular}
}
\end{table}

We perform progressive ablation studies on five real-world traffic datasets with Qwen2.5-3B as the backbone to investigate the effect of four modules: the policy warm-up stage and three core designs: step quality reward (SQ), step importance (SI), and semantic step tree advantage (SSTA). Specifically, SQ introduces averaged step-level quality rewards into final outcome rewards, SI reallocates global advantages across reasoning steps via normalized importance weights, and SSTA builds tree-based semantic advantages to instantiate the full STeP-PO paradigm. Table \ref{tab:ablation} demonstrate consistent performance improvements via progressive module integration. The warm-up stage efficiently helps the LLM grasp basic traffic reasoning capabilities. SQ incentivizes valid reasoning steps, SI prioritizes critical reasoning steps, and SSTA enables fair advantage assignment via a tree-based structure. The full STeP-PO obtains the best performance, validating the synergistic effect of all three modules.

\subsection{Effectiveness of Process Supervision}
\label{sec:rq3_process_supervision}

\begin{table}[t]
\centering
\caption{Effectiveness of process-level supervision.}
\label{tab:process_supervision}
\resizebox{\linewidth}{!}{
\begin{tabular}{lcc}
\toprule
\multirow{2}{*}{Training Method} & \multicolumn{2}{c}{Average Travel Time (s) $\downarrow$} \\
\cmidrule(lr){2-3}
& Jinan 1 & Hangzhou 1\\
\midrule
Warm-up only & 321.23 & 352.12 \\
Warm-up + Outcome-level Reward & 305.74 & 334.25 \\
Warm-up + Step-wise Reward & 293.64 & 321.47 \\
\textbf{Warm-up + STeP-PO} & \textbf{289.34} & \textbf{314.99} \\
\bottomrule
\end{tabular}
}
\end{table}

To verify how process-level supervision provides effective optimization signal for LLM-based traffic signal control, we compare outcome-only RL with our step-wise process optimization. Specifically, we evaluate four settings on Jinan1 and Hangzhou1 using ATT as the metric: (1) \textit{Warm-up only}, where the model is trained only by supervised fine-tuning; (2) \textit{Warm-up + Outcome-level Reward}, where the model is optimized by the final outcome reward only; (3) \textit{Warm-up + Step-wise Reward}, where the model is further guided by the proposed step quality reward; and (4) our \textit{STeP-PO}. In Table~\ref{tab:process_supervision}, both outcome-level reward and process-level supervision improve over the warm-up model, while our STeP-PO performs best. This further demonstrates that step-wise policy optimization is more effective in enhancing LLM reasoning capabilities, as it provides more fine-grained supervision and mitigates the credit misassignment issue.

\subsection{Training Dynamics of ProcessLight}

% \begin{figure}[t]
%       \centering
%       \includegraphics[width=0.48\linewidth]{figures/reward_dynamics_claim_error_count.pdf}
%       \hfill
%       \includegraphics[width=0.48\linewidth]{figures/reward_dynamics_outcome_mean.pdf}
%       \caption{Training dynamics of reward-side indicators. Left: average error claims count. Right: average outcome
%   reward.}
%       \label{fig:reward-dynamics}
% \end{figure}

\begin{figure}[t]
      \centering
      \begin{subfigure}[b]{0.9\linewidth}
          \centering
          \includegraphics[width=\linewidth]{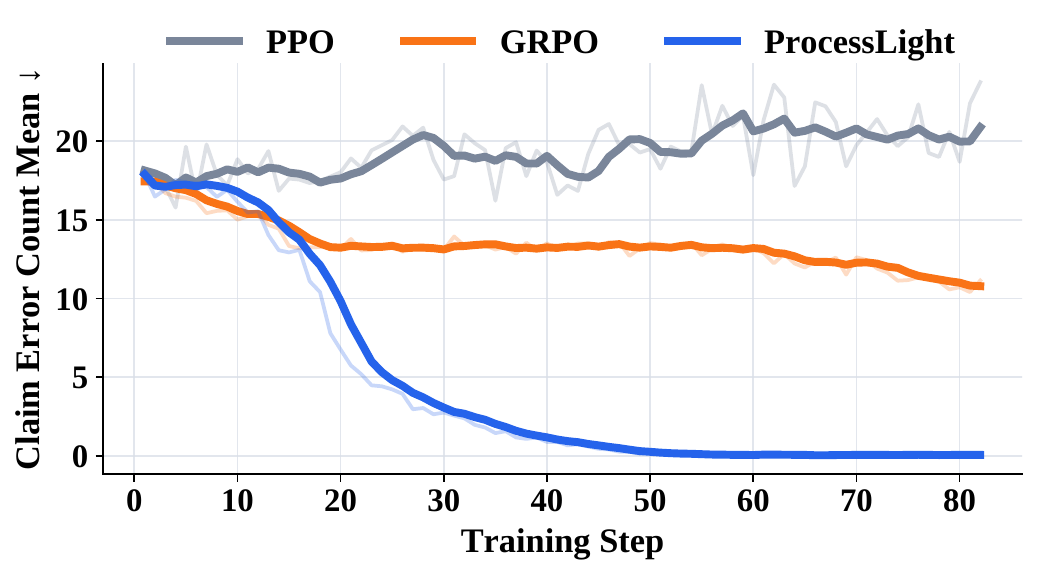}
          \caption{Average error claims count}
          \label{fig:reward-dynamics-a}
      \end{subfigure}

      \begin{subfigure}[b]{0.9\linewidth}
          \centering
          \includegraphics[width=\linewidth]{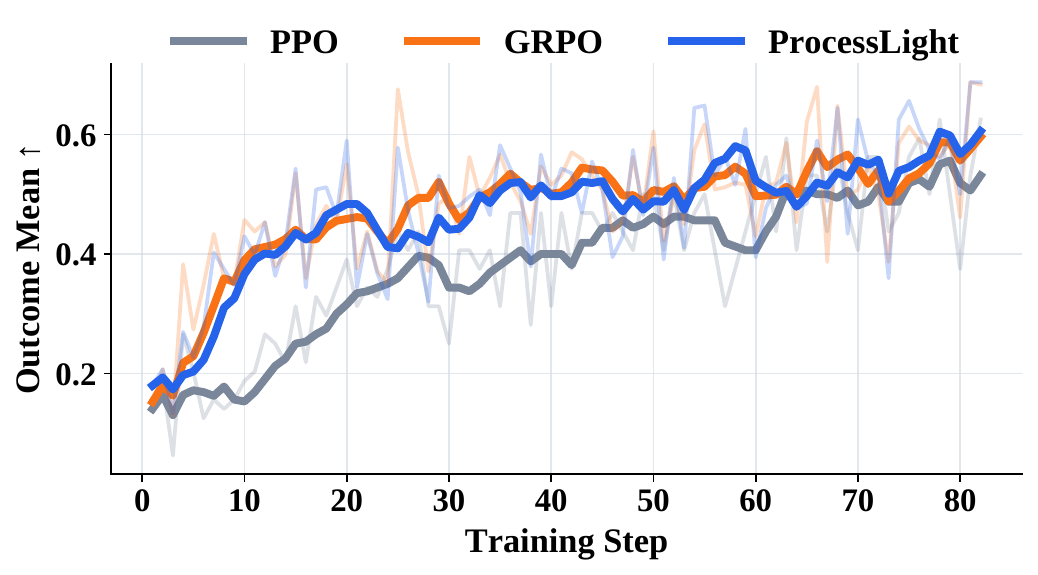}
          \caption{Average outcome reward}
          \label{fig:reward-dynamics-b}
      \end{subfigure}
      
      \caption{Training dynamics of reward-side indicators.}
      \label{fig:reward-dynamics}
\end{figure}

To investigate how the process design of ProcessLight affects RL training, we analyze the reward dynamics of different training methods. Specifically, we compare ProcessLight with PPO and GRPO on two process-side indicators: average error claims count and average outcome reward. The former indicator measures the quality of intermediate reasoning. Lower values indicate fewer incorrect traffic claims. The latter indicator measures the quality of the final phase decision.

As shown in Figure~\ref{fig:reward-dynamics}, ProcessLight rapidly reduces the average error-claim count and consistently maintains a lower error level than PPO and GRPO throughout training. This result shows that process-aware reward verification and step-level credit assignment provide more targeted supervision for intermediate traffic reasoning, leading to more stable and reliable reasoning trajectories. ProcessLight also achieves the best average outcome reward among the compared methods. Over the last 10 training steps, ProcessLight obtains an average outcome reward of 0.585, compared with 0.574 for GRPO and 0.522 for PPO. Thus, improving intermediate reasoning correctness does not compromise final phase selection. Instead, reliable traffic reasoning can support better control decisions.

\subsection{Effectiveness of Step Quality}
\label{app:step_quality}
 Table~\ref{tab:advantage_sign} reports the accuracy of the resulting advantage signs under step-quality and outcome-only supervision. Specifically, by first sampling 100 traffic states to generate reasoning trajectories using Qwen2.5-3B, we then inject two types of controlled noise into each trajectory: logically reversing observation-consistent claims, and replacing the final action with a valid but verified suboptimal phase. Using DeepSeek-V4-Pro as the judge model~\cite{gptscore}, we label the correctness of each reasoning step, and determine the ground-truth sign of the reasoning step advantage. Outcome-only supervision assigns the trajectory-level advantage sign to all reasoning steps. Yet, Step-Quality augments the trajectory-level signal with step-specific validity rewards.

From Table~\ref{tab:advantage_sign}, Outcome-only supervision assigns an incorrect optimization direction to 22\% of the reasoning steps, whereas Step Quality
correctly identifies all injected intermediate errors. This indicates that Step-Quality can distinguish reliable reasoning steps from
hallucinated ones and assign them correct optimization directions.

\begin{table}[t]
\centering
\caption{Accuracy of step-level advantage signs under controlled noise injection.}
\label{tab:advantage_sign}
\begin{tabular}{lc}
\toprule
Supervision & Advantage-Sign Accuracy \\
\midrule
Outcome-only & 78\% \\
Step-Quality & \textbf{100\%} \\
\bottomrule
\end{tabular}
\end{table}

\subsection{Effectiveness of Step Importance}
\label{sec:discussion}

\begin{figure}[t]
\centering
\includegraphics[width=.9\linewidth]{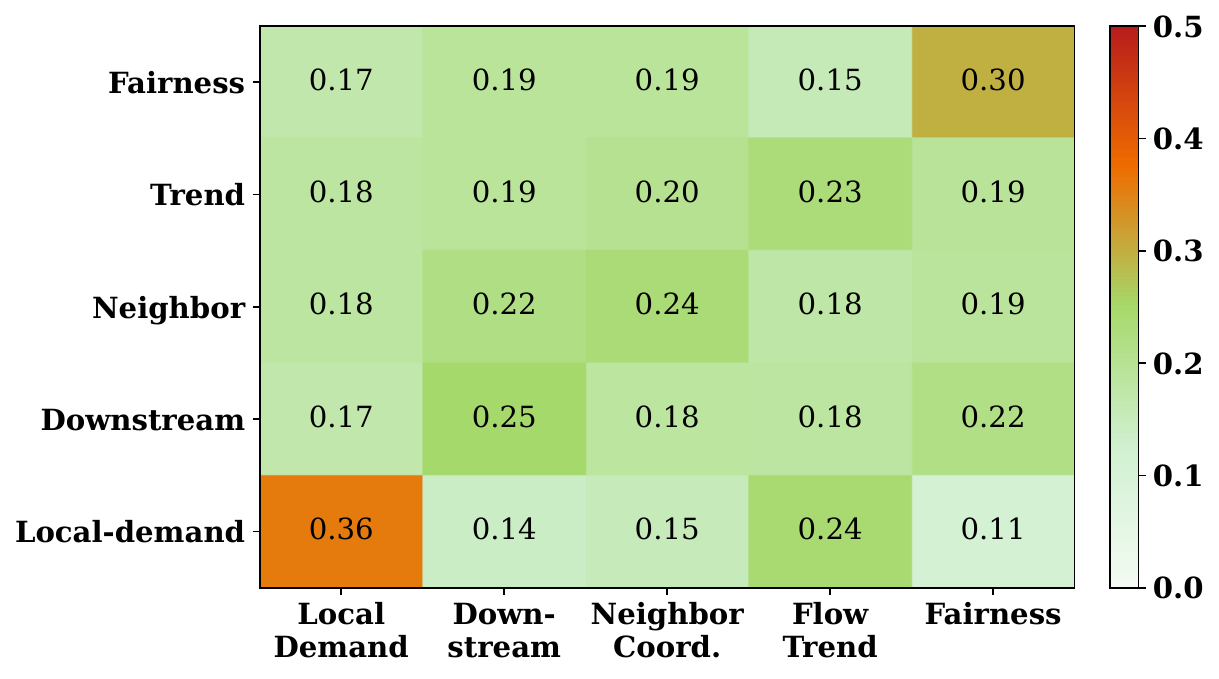}%
\caption{Step importance across traffic regimes. The x-axis denotes reasoning steps, and the y-axis groups traffic regimes by the dominant factor behind the simulator-optimal phase. Each cell shows the normalized importance of a step; higher values indicate a larger change in phase-selection probability after masking that step.}
\label{fig:importance_regime}
\end{figure}

%\noindent\textbf{Effectiveness of Step Importance.}
Finally, to verify {the effectiveness of our step importance method}, we categorize samples into five groups based on the dominant traffic factors that determine the optimal signal phase. We calculate the importance of each reasoning step for all groups and Figure~\ref{fig:importance_regime} plots the visualization heatmap. The cells along the diagonal are significantly brighter, meaning the steps with high importance scores exactly correspond to the core factors affecting traffic efficiency in real scenarios. This result indicates the reliability and effectiveness of our step importance calculation scheme.

\section{Conclusion}
\label{sec:conclusion}

In this paper, we proposed \textbf{ProcessLight}, a process-aware framework for LLM-based traffic signal control. 
Beyond outcome-only policy optimization, it decomposes signal decisions into structured traffic reasoning steps and supervises them through step quality reward, step importance, and semantic step tree advantage. 
Experiments on real-world traffic benchmarks show that ProcessLight leads to better overall performance than traditional traffic control methods, reinforcement learning controllers, and LLM-based baselines. Further evaluations, e.g., ablation studies, confirm that fine-grained reasoning supervision improves both intermediate reasoning reliability and final phase-decision quality.

\section*{Limitations}
Firstly, the claim-based verification incurs high implementation complexity, caused by the structured traffic observations, predefined reasoning steps, and rule-based matching between generated claims and traffic-state evidence. We will explore learnable verifiers for lower efforts. Secondly, step importance estimation introduces extra training overhead, due to counterfactual estimation for reasoning steps. The approximation or amortized estimation strategies could improve ProcessLight. Finally, STeP-PO is designed for structured traffic reasoning with predefined semantic steps on standard four-way intersections. The extension of ProcessLight to flexible free-form  reasoning formats on more complex three- or five-way intersections \cite{shenjj24} remains as future work.

\section*{Acknowledgment}
This work was supported by the National Key R\&D Program of China (Grant No. 2023YFB4301900).

\section*{Ethics Statement}
We abide by the ACL Code of Ethics. The data resources used in this study are publicly available.
During the preparation of this work the authors used ChatGPT in order to polish manuscript language and assist with minor simulation code drafting. After using this tool, the authors reviewed and edited the content as needed and take full responsibility for the content of the published article.

\newpage
\bibliography{custom}

%\clearpage

\appendix

\section{Appendix}
\label{ref:appendix}

\subsection{Concepts in traffic signal control}
\label{app:concepts}

\begin{figure}
    \centering
    \includegraphics[width=1.0\linewidth]{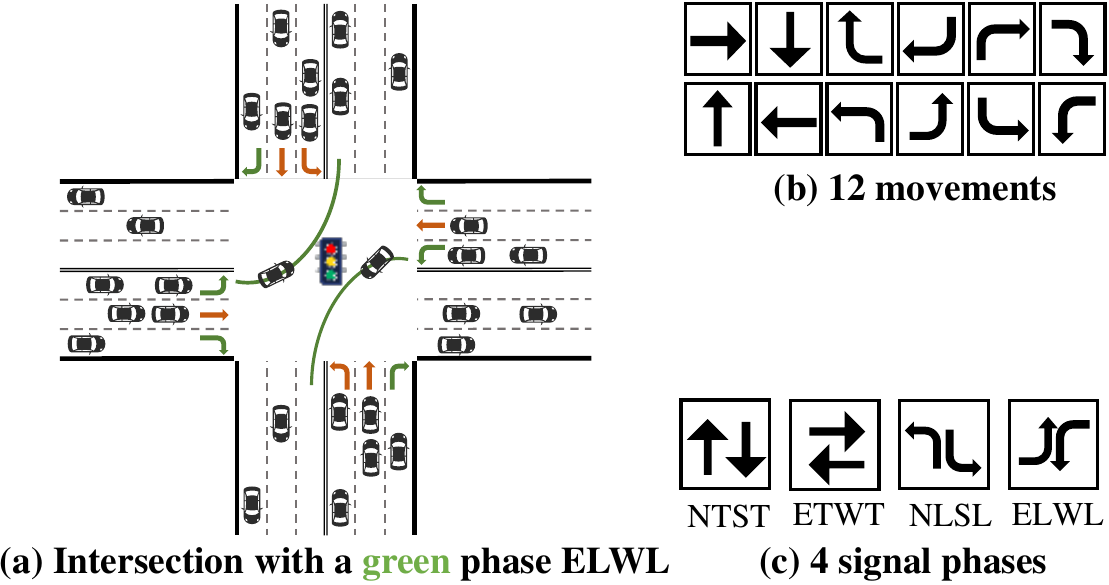}
    \caption{Illustration of (a) intersections, (b) traffic movements, and (c) signal phases.}
    \label{fig:tsc_concepts}
\end{figure}

We illustrate the commonly used  settings in traffic signal control in Figure~\ref{fig:tsc_concepts}.

\subsection{Baseline Methods}
\label{app:baselines}

We categorize the baselines into traditional methods, reinforcement learning methods, and large language model methods.

\subsubsection{Traditional Methods.}
\begin{itemize}
    \item \textbf{FixedTime} \cite{traffictimingmanual}: FixedTime uses predefined cycle lengths, phase orders, and green durations, providing a simple and stable controller that does not adapt to real-time traffic demand.
    
    \item \textbf{MaxPressure} \cite{maxpressure}: MaxPressure selects the phase with the largest pressure difference between upstream queues and downstream traffic conditions, achieving strong queue dissipation through transportation-theoretic control.
\end{itemize}

\subsubsection{RL-based Methods.}
\begin{itemize}
    \item \textbf{PressLight} \cite{presslight}: PressLight applies deep reinforcement learning to optimize intersection pressure, combining adaptive signal control with a congestion-oriented traffic objective.
    
    \item \textbf{MPLight} \cite{mplight}: MPLight builds on FRAP and uses pressure as both observation and reward, enabling pressure-aware phase competition modeling for traffic signal control.
    
    \item \textbf{AttendLight} \cite{attendlight}: AttendLight employs attention mechanisms to construct traffic observations and model phase transition dependencies, allowing the agent to focus on critical movements.
    
    \item \textbf{CoLight} \cite{colight}: CoLight uses graph attention networks for communication among neighboring intersections, enabling cooperative multi-agent signal control over road networks.
    
\item \textbf{Efficient-CoLight} \cite{efficientpressure}: Efficient-CoLight improves CoLight by introducing efficient pressure observations, reducing redundant state information while preserving key coordination signals.
    
    \item \textbf{Advanced-CoLight} \cite{expression}: Advanced-CoLight extends CoLight with richer traffic state features such as pressure and effective running vehicles, improving expressive multi-intersection coordination.
\end{itemize}

\subsubsection{LLM-based Methods.}
\begin{itemize}
    \item \textbf{LLMLight} \cite{llmlight}: LLMLight uses an LLM-based traffic signal control agent to interpret traffic states and generate signal actions through natural-language traffic reasoning.
    
    \item \textbf{Traffic-R1} \cite{trafficr1}: Traffic-R1 develops an LLM-based traffic signal control agent with a two-stage reinforcement learning framework to improve real-world deployment capability.
    
    \item \textbf{CoLLMLight} \cite{collmlight}: CoLLMLight coordinates multiple LLM-based traffic signal control agents with cost-aware reasoning depth, balancing cooperative decision quality and inference overhead.
\end{itemize}

\subsection{Metrics Details}
\label{app:metrics}

\textbf{Average Travel Time (ATT).} 
ATT measures the average time required for vehicles to complete their trips within the road network. It directly reflects the overall traffic efficiency under the learned signal control policy. Formally, ATT is defined as
\begin{equation}
\mathrm{ATT}
=
\frac{1}{N}
\sum_{v=1}^{N}
\left(
t_v^{\mathrm{arrive}}
-
t_v^{\mathrm{depart}}
\right),
\end{equation}
where $N$ denotes the total number of vehicles, and $t_v^{\mathrm{depart}}$ and $t_v^{\mathrm{arrive}}$ denote the departure and arrival times of vehicle $v$, respectively.

\textbf{Average Waiting Time (AWT).}
AWT measures the average accumulated waiting time of vehicles during the trip. A vehicle is considered waiting when its speed falls below a predefined threshold. Lower AWT indicates reduced intersection delay and smoother traffic flow. It is computed as
\begin{equation}
\mathrm{AWT}
=
\frac{1}{N}
\sum_{v=1}^{N}
w_v,
\end{equation}
where $w_v$ is the total waiting time of vehicle $v$.

\textbf{Average Queue Length (AQL).}
AQL measures the average number of queuing vehicles on all lanes and decision steps, indicating the congestion level of the traffic network. It is defined as
\begin{equation}
\mathrm{AQL}
=
\frac{1}{T|\mathcal{L}|}
\sum_{t=1}^{T}
\sum_{l \in \mathcal{L}}
q_l^t,
\end{equation}
where $T$ denotes the total number of simulation steps, $\mathcal{L}$ denotes the set of lanes, and $q_l^t$ represents the queue length of lane $l$ at simulation step $t$.

\subsection{Datasets}
\label{app:datasets}

%\textcolor{red}
All intersections in the Jinan and Hangzhou datasets adopt four-way topology.

\begin{itemize}
    \item \textbf{Jinan}: A dataset from the Dongfeng sub-district in Jinan, China, consisting of 12 intersections with three traffic flow datasets. Every intersection is composed of two 400-meter east-west roads and two 800-meter north-south roads.
    
    \item \textbf{Hangzhou}: A dataset from the Gudang sub-district in Hangzhou, China, consisting of 16 intersections with two traffic flow datasets. Every intersection is composed of two 800-meter east-west roads and two 600-meter north-south roads.

    \item \textbf{Synthetic-1}: A synthetic dataset consisting of 16 intersections and each road between two intersections is 300 meters long.
\end{itemize}

% \begin{table}[t]
% \renewcommand{\arraystretch}{1.1}
%  \small
%  \centering
%  \caption{Statistics of datasets.}
%  \vspace{-5pt}
%  % \setlength{\tabcolsep}{5pt}
%  \resizebox{0.48\textwidth}{!}{
%   \begin{tabular}{c|c|c|cccc}
%    \toprule
%    \multirow{2}{*}{Flow dataset} & \multirow{2}{*}{Structure} & \multirow{2}{*}{Vehicles} & \multicolumn{4}{c}{Arrival rate (vehicles/5min)} \\
%    \cline{4-7}
%    &  & & Mean & Std & Max & Min \\
%    \midrule
%    Jinan 1 & \multirow{3}{*}{3$\times$4} & 6295 & 523.67 & 98.52 & 671 & 255 \\
%    Jinan 2 & & 4365 & 362.83 & 74.81 & 493 & 236 \\
%    Jinan 3 & & 5494 & 456.92 & 46.38 & 544 & 362 \\
%    \midrule
%    Hangzhou 1 & \multirow{2}{*}{4$\times$4} & 2983 & 247.67 & 40.44 & 332 & 211\\
%    Hangzhou 2 & & 6984 & 581.08 & 318.43 & 1145 & 202 \\
%    \midrule
%    Synthetic 1 & {4$\times$4} & 8000 & 666.6 & 32.10 & 735 & 612 \\
%    \bottomrule
%   \end{tabular}}
%   \label{tab:dataStats}
%   % \vspace{-10pt}
% \end{table}

\subsection{STeP-PO Algorithm}
\label{app:alg}

As shown in Algorithm~\ref{alg:step_po}, STeP-PO consists of three sequential stages: training data collection, policy warm-up, and step-wise policy optimization.

\begin{algorithm}[t]
\caption{STeP-PO}
\label{alg:step_po}
\begin{algorithmic}[1]
\Require Policy model $\pi_\theta$ initialized by a pre-trained LLM; teacher model $\pi_{\mathrm{tea}}$; traffic simulator $\mathcal{S}$; traffic scenarios $\mathcal{C}$; group size $G$
\Ensure Trained traffic signal control policy $\pi_\theta$

\Statex \textcolor{gray}{\textit{Training data collection:}}
\State Collect traffic scenarios from $\mathcal{S}$ and construct input prompts following Section~\ref{subsec:processlight_framework}
\State Select oracle phase actions through network-level simulation and collect teacher-generated step-wise reasoning trajectories
\State Build dataset $\mathcal{D}=\{(x,y)\}$ and split it into $\mathcal{D}_{\mathrm{SFT}}$ and $\mathcal{D}_{\mathrm{RL}}$

\Statex \textcolor{gray}{\textit{Policy warm-up:}}
\For{each sample $(x,y) \in \mathcal{D}_{\mathrm{SFT}}$}
    \State Optimize $\pi_\theta$ with the supervised warm-up objective in Eq.~\ref{eq:warmup_loss}
\EndFor
\State Set reference policy $\pi_{\mathrm{ref}} \leftarrow \pi_\theta$

\Statex \textcolor{gray}{\textit{Step-wise policy optimization:}}
\For{each sample $(x,y^\star) \in \mathcal{D}_{\mathrm{RL}}$}
    \State Generate a group of reasoning trajectories $\{y_i\}_{i=1}^{G} \sim \pi_\theta(\cdot|x)$
    \State Compute step quality rewards using the reference trajectory $y^\star$ by Eq.~\ref{eq:step_reward}
    \State Estimate phase-sensitive step importance by Eq.~\ref{eq:step_imp}
    \State Construct the semantic step tree and compute step advantages by Eq.~\ref{eq:step_adv}
    \State Optimize $\pi_\theta$ with the STeP-PO objective in Eq.~\ref{eq:step_po_obj}
\EndFor

\State \Return $\pi_\theta$
\end{algorithmic}
\end{algorithm}

\subsection{Semantic Step Tree Construction}
\label{app:tree_construction}

We present the semantic step tree construction algorithm in Algorithm~\ref{alg:semantic_step_tree}.

\begin{algorithm}[t]
\caption{Semantic Step Tree Construction}
\label{alg:semantic_step_tree}
\begin{algorithmic}[1]
\Require Input prompt $x$; sampled trajectories $\{y_i\}_{i=1}^{G}$, where $y_i=(s_i^1,\ldots,s_i^K)$; final action of each trajectory $a_i$.
\Ensure Semantic step tree $\mathcal{T}$.

\State Initialize $\mathcal{T}$ with a root node $n_0$ representing the input prompt $x$.
\For{$i=1$ to $G$}
    \State Set the current node $n \leftarrow n_0$.
    \For{$k=1$ to $K$}
        \State Extract the claim set $C_i^k$ from reasoning step $s_i^k$.
        \State Search among the children of $n$ for a node $n'$ at step $k$ whose claim set is semantically matched with $C_i^k$.
        \If{such a node $n'$ exists}
            \State Merge $s_i^k$ into $n'$ and add it to the step set $\mathcal{S}(n')$.
            \State Set $n \leftarrow n'$.
        \Else
            \State Create a new child node $n'$ under $n$.
            \State Store $C_i^k$, $s_i^k$, and its step index $k$ in $n'$.
            \State Set $\mathcal{S}(n')=\{s_i^k\}$ and $n \leftarrow n'$.
        \EndIf
    \EndFor
    \State Attach the final action $a_i$ as a leaf node under $n$.
\EndFor
\State \Return $\mathcal{T}$.
\end{algorithmic}
\end{algorithm}

\subsection{Hyperparameters Analysis}

\begin{table}[t]
\centering
\small
\caption{Hyperparameter sensitivity of STeP-PO on Jinan1 and Hangzhou1.}
\label{tab:hyperparam}
\resizebox{\linewidth}{!}{
\begin{tabular}{cccccc}
\toprule
\# & $\eta$ & $\lambda$ & $G$ & Jinan1 ATT $\downarrow$ & Hangzhou1 ATT $\downarrow$ \\
\midrule
1 & 0.3 & 0.2 & 4  & 307.87 & 337.24 \\
2 & 0.3 & 0.3 & 8  & 291.62 & 319.33 \\
3 & \textbf{0.6} & \textbf{0.3} & \textbf{8}  & \textbf{289.34} & \textbf{314.99} \\
4 & 0.6 & 0.4 & 8  & 296.76 & 320.52 \\
5 & 0.9 & 0.3 & 12 & 294.48 & 321.05 \\
6 & 0.9 & 0.5 & 12 & 302.15 & 329.41 \\
\bottomrule
\end{tabular}
}
\end{table}

We analyze the sensitivity of STeP-PO to three key hyperparameters on Jinan1 and Hangzhou1 using ATT as the evaluation metric. Specifically, $\eta$ controls the strength of phase-sensitive step importance weighting, $\lambda$ balances the step-level advantage and outcome-level feedback, and $G$ denotes the number of sampled reasoning trajectories used to construct the semantic step tree. As shown in Table~\ref{tab:hyperparam}, STeP-PO achieves the best performance with $\eta=0.6$, $\lambda=0.3$, and $G=8$, which is used as the default setting in our main experiments. Increasing $G$ from 4 to 8 consistently reduces ATT on both datasets, indicating that more rollouts improve reasoning diversity and provide richer sibling branches for more stable advantage estimation. 

The results also show that both $\lambda$ and $\eta$ should be set to moderate values. A proper $\lambda$ improves performance by incorporating step-quality supervision into policy optimization, while a larger value degrades ATT because overemphasizing step-level rewards weakens the supervision from the final phase outcome. Similarly, increasing $\eta$ from 0.3 to 0.6 improves ATT, demonstrating that reasonable importance weighting helps optimize decision-critical reasoning steps. Nevertheless, when $\eta$ is further increased to 0.9, performance declines, implying that excessive reliance on a few highly sensitive steps may ignore the contribution of the complete reasoning trajectory. These results confirm that STeP-PO benefits from a balanced combination of rollout diversity, step-quality supervision, and phase-sensitive credit allocation.

\subsection{Statistical Significance Analysis}
\label{app:significance}

Each model is evaluated using five random seeds (11, 42, 1024, 2026, 10001) on Jinan1 and Hangzhou1.
Table~\ref{tab:significance} reports the mean and standard deviation across
seeds. ProcessLight achieves the best mean performance on most metrics, and
its coefficient of variation ranges from 0.44\% to 9.76\%, indicating limited
sensitivity to random-seed variation.

To further assess robustness, we conduct two-sided Welch's t-tests comparing
each baseline against ProcessLight and apply one global Holm correction
across all 90 comparisons. ProcessLight obtains a lower mean in 88 of the 90
comparisons; 81 comparisons significantly favor ProcessLight after
correction, 9 show no detectable difference, and none significantly favors a
baseline. The nine non-significant comparisons involve AdvancedCoLight on
Jinan1 (ATT) and Hangzhou1 (AWT/AQL), LLMLight on Jinan1 (ATT/AWT) and
Hangzhou1 (AWT), and Traffic-R1 on Jinan1 (ATT/AWT) and Hangzhou1 (AWT).
Although AdvancedCoLight achieves lower mean AWT and AQL on Hangzhou1, neither
difference remains statistically significant after Holm correction.

\begin{table*}[t]
\centering
\caption{Mean and standard deviation over five random seeds on Jinan1 and Hangzhou1. Table~\ref{tab:main_results} reports results under the default seed. Best results are in bold.}
\label{tab:significance}
\resizebox{\textwidth}{!}{
\begin{tabular}{lcccccc}
\toprule
\multirow{2}{*}{Method} & \multicolumn{3}{c|}{Jinan 1} & \multicolumn{3}{c}{Hangzhou 1} \\
\cmidrule(lr){2-4}\cmidrule(lr){5-7}
& ATT$\downarrow$ & AWT$\downarrow$ & AQL$\downarrow$ & ATT$\downarrow$ & AWT$\downarrow$ & AQL$\downarrow$ \\
\midrule
\multicolumn{7}{c}{\textbf{Traditional Methods}} \\
\midrule
FixedTime & 409.73 $\pm$ 0.00 & 156.76 $\pm$ 0.00 & 190.07 $\pm$ 0.00 & 564.86 $\pm$ 0.00 & 204.63 $\pm$ 0.00 & 169.56 $\pm$ 0.00 \\
MaxPressure & 305.26 $\pm$ 0.00 & 60.19 $\pm$ 0.00 & 72.97 $\pm$ 0.00 & 343.94 $\pm$ 0.00 & 47.09 $\pm$ 0.00 & 39.02 $\pm$ 0.00 \\
\midrule
\multicolumn{7}{c}{\textbf{RL Methods}} \\
\midrule
EfficientPressLight & 316.85 $\pm$ 0.00 & 77.93 $\pm$ 0.00 & 136.27 $\pm$ 0.00 & 472.22 $\pm$ 0.01 & 178.99 $\pm$ 0.00 & 148.31 $\pm$ 0.00 \\
EfficientCoLight & 310.22 $\pm$ 0.01 & 70.10 $\pm$ 0.00 & 122.57 $\pm$ 0.00 & 340.52 $\pm$ 0.00 & 43.85 $\pm$ 0.00 & 36.34 $\pm$ 0.00 \\
AttendLight & 514.56 $\pm$ 21.96 & 251.71 $\pm$ 16.56 & 440.15 $\pm$ 28.95 & 481.44 $\pm$ 6.99 & 178.86 $\pm$ 7.56 & 148.21 $\pm$ 6.26 \\
EfficientMPLight & 311.36 $\pm$ 0.07 & 71.33 $\pm$ 0.14 & 124.72 $\pm$ 0.26 & 384.46 $\pm$ 0.01 & 93.46 $\pm$ 0.00 & 77.44 $\pm$ 0.00 \\
AdvancedCoLight & 298.81 $\pm$ 0.00 & 60.40 $\pm$ 0.00 & 105.61 $\pm$ 0.00 & 328.58 $\pm$ 0.00 & \textbf{32.95 $\pm$ 0.00} & \textbf{27.30 $\pm$ 0.00} \\
MPLight & 489.20 $\pm$ 15.31 & 233.69 $\pm$ 13.35 & 408.62 $\pm$ 23.34 & 517.53 $\pm$ 15.94 & 208.65 $\pm$ 13.73 & 172.89 $\pm$ 11.38 \\
AdvancedMPLight & 313.15 $\pm$ 0.00 & 75.18 $\pm$ 0.00 & 131.47 $\pm$ 0.00 & 372.21 $\pm$ 0.01 & 80.23 $\pm$ 0.00 & 66.48 $\pm$ 0.00 \\
CoLight & 474.29 $\pm$ 21.28 & 215.33 $\pm$ 16.43 & 376.53 $\pm$ 28.73 & 485.46 $\pm$ 9.60 & 175.46 $\pm$ 6.92 & 145.38 $\pm$ 5.73 \\
\midrule
\multicolumn{7}{c}{\textbf{LLM-based Methods}} \\
\midrule
LightGPT-3B & 356.32 $\pm$ 4.60 & 110.68 $\pm$ 4.10 & 134.21 $\pm$ 4.97 & 378.87 $\pm$ 3.23 & 82.81 $\pm$ 2.49 & 68.62 $\pm$ 2.07 \\
LightGPT-8B & 337.62 $\pm$ 1.15 & 94.79 $\pm$ 1.16 & 114.93 $\pm$ 1.41 & 395.63 $\pm$ 2.85 & 105.55 $\pm$ 2.84 & 87.45 $\pm$ 2.35 \\
LLMLight & 297.58 $\pm$ 0.24 & 52.69 $\pm$ 0.29 & 63.89 $\pm$ 0.35 & 337.04 $\pm$ 0.53 & 41.38 $\pm$ 0.37 & 34.29 $\pm$ 0.31 \\
CoLLMLight & 303.56 $\pm$ 1.14 & 59.05 $\pm$ 1.11 & 71.61 $\pm$ 1.35 & 342.71 $\pm$ 1.09 & 47.86 $\pm$ 0.90 & 39.66 $\pm$ 0.75 \\
Traffic-R1 & 299.13 $\pm$ 0.45 & 54.01 $\pm$ 0.40 & 65.49 $\pm$ 0.49 & 336.56 $\pm$ 0.82 & 40.04 $\pm$ 0.73 & 33.17 $\pm$ 0.61 \\
\midrule
\multicolumn{7}{c}{\textbf{ProcessLight (ours)}} \\
\midrule
Qwen2.5-3B & 335.26 $\pm$ 3.74 & 90.49 $\pm$ 3.69 & 109.72 $\pm$ 4.48 & 359.00 $\pm$ 1.20 & 64.67 $\pm$ 1.60 & 53.59 $\pm$ 1.32 \\
Qwen3-8B & 308.79 $\pm$ 5.49 & 74.75 $\pm$ 2.74 & 83.14 $\pm$ 2.04 & 342.45 $\pm$ 7.88 & 45.53 $\pm$ 3.45 & 37.16 $\pm$ 2.53 \\
Qwen3-32B & 310.31 $\pm$ 1.28 & 64.20 $\pm$ 1.17 & 77.84 $\pm$ 1.42 & 339.53 $\pm$ 0.95 & 42.52 $\pm$ 0.80 & 35.23 $\pm$ 0.67 \\
DeepSeek-V4-Flash & 297.53 $\pm$ 5.09 & 53.92 $\pm$ 4.35 & 63.61 $\pm$ 3.79 & 323.84 $\pm$ 3.43 & 45.77 $\pm$ 4.49 & 35.43 $\pm$ 3.52 \\
DeepSeek-V4-Pro & 291.31 $\pm$ 2.74 & 54.65 $\pm$ 2.84 & 63.62 $\pm$ 2.87 & 325.56 $\pm$ 3.50 & 43.21 $\pm$ 2.75 & 35.30 $\pm$ 3.38 \\
ProcessLight & \textbf{290.55 $\pm$ 4.50} & \textbf{46.31 $\pm$ 4.52} & \textbf{53.13 $\pm$ 1.17} & \textbf{312.61 $\pm$ 1.38} & 35.66 $\pm$ 3.27 & 29.27 $\pm$ 1.14 \\
\bottomrule
\end{tabular}
}
\end{table*}

\if 0
\subsection{Effectiveness of Step Quality}
\label{app:step_quality}

To directly compare step-quality supervision with outcome-only supervision, we
sample 100 traffic states and generate reasoning trajectories using
Qwen2.5-3B. We then inject two types of controlled noise into each trajectory:
logically reversing observation-consistent claims, and replacing the final
action with a valid but verified suboptimal phase. DeepSeek-V4-Pro is used as
the judge model~\cite{gptscore} to label the correctness of each reasoning
step, which determines the ground-truth sign of its advantage. Outcome-only
supervision assigns the trajectory-level advantage sign to all reasoning
steps, whereas Step Quality augments the trajectory-level signal with
step-specific validity rewards.

Table~\ref{tab:advantage_sign} reports the accuracy of the resulting advantage
signs under both methods. Outcome-only supervision assigns an incorrect
optimization direction to 22\% of the reasoning steps, whereas Step Quality
correctly identifies all injected intermediate errors. This indicates that
Step Quality enables STeP-PO to distinguish reliable reasoning steps from
hallucinated ones and assign them correct optimization directions.

\begin{table}[t]
\centering
\caption{Accuracy of step-level advantage signs under controlled noise injection.}
\label{tab:advantage_sign}
\begin{tabular}{lc}
\toprule
Supervision & Advantage-Sign Accuracy \\
\midrule
Outcome-only & 78\% \\
With Step Quality & \textbf{100\%} \\
\bottomrule
\end{tabular}
\end{table}
\fi 

\subsection{Reliability of Claim Verification}
\label{app:verification}

The step-quality evaluation pipeline consists of three stages:
regular-expression parsing of the generated claims, soft semantic matching
against the reference key claims~\cite{r1vl}, and observation-based validity
checking. To quantify the reliability of each stage, we randomly sample 100
instances, roll out 8 responses for each prompt, and conduct three independent
evaluations using DeepSeek-V4-Pro as the judge model~\cite{distjudge}. As
shown in Table~\ref{tab:verification_accuracy}, all three stages achieve high
execution accuracy, supporting the reliability of the step-quality signal
for policy optimization.

\begin{table}
\centering
\caption{Execution accuracy of the three verification stages.}
\label{tab:verification_accuracy}
\begin{tabular}{lc}
\toprule
Component & Execution Accuracy \\
\midrule
Claim extraction & 99.81 $\pm$ 0.02\% \\
Semantic matching & 96.87 $\pm$ 0.04\% \\
Observation validity & 97.25 $\pm$ 0.46\% \\
\bottomrule
\end{tabular}
\end{table}

\setcounter{dbltopnumber}{3}
\begin{table*}
\centering
\caption{Cost of the two data-construction stages.}
\label{tab:cost}
\resizebox{\linewidth}{!}{
\begin{tabular}{lccccc}
\toprule
Stage & Model / Environment & Calls & Token Usage & Runtime & Cost \\
\midrule
Reference claims & DeepSeek-V4-Pro, Cloud API & 3k & 7.07M in + 1.70M out & -- & \$9.1 \\
Step importance & Qwen2.5-3B, 1$\times$ RTX PRO 6000 & 21k & 57.81M in; no output & 1.5 GPU-h & \$1.2 \\
\bottomrule
\end{tabular}
}
\end{table*}

\begin{table*}[t]
\centering
\caption{Zero-shot performance under restricted observation without neighboring-intersection information. Best results are in bold.}
\label{tab:restricted}
\resizebox{\textwidth}{!}{
\begin{tabular}{lccccccccccccccc}
\toprule
\multirow{2}{*}{Method} & \multicolumn{3}{c|}{Jinan 1} & \multicolumn{3}{c|}{Jinan 2} & \multicolumn{3}{c|}{Jinan 3} & \multicolumn{3}{c|}{Hangzhou 1} & \multicolumn{3}{c}{Hangzhou 2} \\
\cmidrule(lr){2-4}\cmidrule(lr){5-7}\cmidrule(lr){8-10}\cmidrule(lr){11-13}\cmidrule(lr){14-16}
& ATT$\downarrow$ & AWT$\downarrow$ & AQL$\downarrow$ & ATT$\downarrow$ & AWT$\downarrow$ & AQL$\downarrow$ & ATT$\downarrow$ & AWT$\downarrow$ & AQL$\downarrow$ & ATT$\downarrow$ & AWT$\downarrow$ & AQL$\downarrow$ & ATT$\downarrow$ & AWT$\downarrow$ & AQL$\downarrow$ \\
\midrule
LLMLight & \textbf{297.28} & 52.28 & \textbf{63.39} & 310.48 & 70.66 & 123.56 & \textbf{291.86} & \textbf{55.96} & 85.40 & 337.21 & 41.60 & 34.47 & 419.60 & 84.91 & \textbf{164.73} \\
ProcessLight w/o Neighbor Info. & 300.91 & \textbf{48.87} & 65.63 & \textbf{304.32} & \textbf{68.35} & \textbf{119.47} & 299.48 & 56.23 & \textbf{81.59} & \textbf{332.06} & \textbf{39.14} & \textbf{34.38} & \textbf{414.50} & \textbf{83.65} & 168.67 \\
\bottomrule
\end{tabular}
}
\end{table*}

\begin{table*}
\centering
\caption{Zero-shot performance with different backbones and teacher models. Best results are in bold.}
\label{tab:backbone}
\resizebox{\textwidth}{!}{
\begin{tabular}{llccccccccccccccc}
\toprule
\multirow{2}{*}{Method} & \multirow{2}{*}{Teacher} & \multicolumn{3}{c|}{Jinan 1} & \multicolumn{3}{c|}{Jinan 2} & \multicolumn{3}{c|}{Jinan 3} & \multicolumn{3}{c|}{Hangzhou 1} & \multicolumn{3}{c}{Hangzhou 2} \\
\cmidrule(lr){3-5}\cmidrule(lr){6-8}\cmidrule(lr){9-11}\cmidrule(lr){12-14}\cmidrule(lr){15-17}
& & ATT$\downarrow$ & AWT$\downarrow$ & AQL$\downarrow$ & ATT$\downarrow$ & AWT$\downarrow$ & AQL$\downarrow$ & ATT$\downarrow$ & AWT$\downarrow$ & AQL$\downarrow$ & ATT$\downarrow$ & AWT$\downarrow$ & AQL$\downarrow$ & ATT$\downarrow$ & AWT$\downarrow$ & AQL$\downarrow$ \\
\midrule
LLMLight & -- & 297.28 & 52.28 & 63.39 & 310.48 & 70.66 & 123.56 & 291.86 & 55.96 & 85.40 & 337.21 & 41.60 & 34.47 & 419.60 & 84.91 & 164.73 \\
CoLLMLight & -- & 298.99 & 54.35 & 65.90 & 311.23 & 71.06 & 124.25 & 292.25 & 56.58 & 86.35 & 336.25 & 41.30 & 34.22 & 419.96 & 74.02 & \textbf{143.59} \\
Traffic-R1 & -- & 299.14 & 54.15 & 65.65 & 310.79 & 70.54 & 123.34 & 291.96 & 56.06 & 85.56 & 337.46 & 40.94 & 33.92 & 421.28 & 81.03 & 157.20 \\
\midrule
ProcessLight (Qwen2.5-3B) & DeepSeek-V4-Pro & 289.34 & 45.03 & 55.94 & \textbf{293.18} & \textbf{54.16} & 102.67 & \textbf{284.33} & 50.97 & 78.20 & 314.99 & 32.82 & 30.48 & 407.24 & 73.13 & 154.21 \\
ProcessLight (Qwen3-8B) & DeepSeek-V4-Pro & \textbf{288.81} & \textbf{43.18} & 54.49 & 295.60 & 54.57 & 100.39 & 285.15 & \textbf{48.43} & \textbf{77.18} & 306.95 & \textbf{30.98} & \textbf{30.13} & \textbf{401.47} & \textbf{71.06} & \textbf{152.25} \\
ProcessLight (Llama3-8B) & DeepSeek-V4-Pro & 289.98 & 46.35 & \textbf{52.57} & 299.28 & 63.17 & \textbf{96.90} & 294.29 & 55.03 & 82.13 & \textbf{306.15} & 35.29 & 38.96 & 409.45 & 72.54 & 156.53 \\
ProcessLight (Qwen2.5-3B) & Qwen3-Plus & 290.73 & 47.66 & 56.29 & 296.88 & 67.42 & 105.32 & 301.74 & 56.50 & 87.26 & 309.72 & 37.05 & 45.56 & 410.15 & 79.60 & 152.62 \\
\bottomrule
\end{tabular}
}
\end{table*}

\subsection{Generalization across Backbones and Teachers}
\label{app:backbone}

We further evaluate ProcessLight with Qwen3-8B and Llama3-8B as backbone
models, and with Qwen3-Plus as the teacher model for training-data collection.
Table~\ref{tab:backbone} reports the zero-shot performance across the five
real-world datasets. ProcessLight remains effective across the tested
settings: all three backbone variants outperform LLMLight, CoLLMLight, and
Traffic-R1 on most metrics, and the Qwen3-8B variant improves over the
Qwen2.5-3B variant on 12 of the 15 metrics. These results provide initial
evidence that the method is not tied to a single backbone family or to the
original teacher model; our conclusions are delimited to the models evaluated
here.

\subsection{Restricted-Observation Comparison}
\label{app:restricted}

To isolate the contribution of the training strategy from the richer
observation, we remove neighboring-intersection information and the
corresponding reasoning guidance from the ProcessLight input, and evaluate
the modified model on all five datasets. As shown in
Table~\ref{tab:restricted}, the restricted variant still outperforms LLMLight
on most metrics. We note that the restricted-observation experiment serves as
a fairness control that isolates the contribution of the training strategy,
whereas the full model reflects the intended network-coordination problem
setting.

\subsection{Computational Cost Analysis}
\label{app:cost}

Prior to RL optimization, we precompute the step-importance scores for each
prompt and reuse them across all rollouts associated with that prompt;
compared with GRPO, step-importance estimation therefore introduces no
additional model forward or backward passes during training. For $N$ prompts
with $K$ reasoning steps, generating the reference claims requires $N$ calls
to the teacher model, while step-importance estimation requires $N(K+1)$ calls
to the SFT-initialized policy model---one for the complete response and $K$
for the step-masked variants. Step-importance estimation thus accounts for
$\frac{K+1}{K+2}$ of all model calls, i.e., $87.5\%$ with $K=6$. This
call-count proportion overestimates the practical overhead, because these
calls use a locally deployed 3B model for forward-only scoring, without
autoregressive decoding or additional traffic simulation. As shown in
Table~\ref{tab:cost}, step-importance estimation accounts for only 11.6\% of
the combined monetary cost of the data-construction pipeline.

\subsection{Prompt Template}
\label{app:prompt}

\onecolumn
\begin{table*}[t]
    \centering
    \small
    \begin{lstlisting}
You are an expert traffic signal control agent operating under a strict reasoning framework.
Your task is to select exactly one legal phase for the current intersection based on the provided traffic facts.

## Background Context
You are a traffic signal controller at a four-way intersection with 12 movements:
[NL, NT, NR, SL, ST, SR, EL, ET, ER, WL, WT, WR] and 4 phases
[ETWT, NTST, ELWL, NLSL]. Each phase releases two compatible movements:
ETWT releases ET and WT, NTST releases NT and ST, ELWL releases EL and WL,
and NLSL releases NL and SL.

Legal phases for this decision: [ETWT, NTST, ELWL, NLSL].

## Decision Rules
1. Prioritize local intersection pressure relief; grant right-of-way first to long queues and stop-line nearby vehicles.
2. Assess downstream congestion before releasing any movement. Avoid discharging vehicles into congested downstream sections.
3. Use adjacent intersections only as local risk reference in this decision step.
4. Avoid prolonged single-lane waiting when downstream conditions permit.
5. Occupancy 0-0.2 indicates free flow, 0.2-0.4 basically smooth, above 0.4 mild congestion, and above 0.6 severe congestion.

## Current Intersection State

### Movement State

| Movement | Phase | Early Queue Length | Downstream Queue Length | Pressure | Avg Waiting Time(minutes) | Segment Veh Occupancy | Vehicle Count |
| --- | --- | --- | --- | --- | --- | --- | --- |
| ET | ETWT | 0 | 10 | -3.3 | 0.1 | [0, 0.2, 0] | 3 |
| WT | ETWT | 0 | 2 | -0.7 | 0 | [0, 0.1, 0] | 2 |
| NT | NTST | 1 | 0 | 1 | 0.1 | [0, 0.1, 0.1] | 9 |
| ST | NTST | 1 | 5 | -0.7 | 0 | [0.1, 0, 0.1] | 10 |
| EL | ELWL | 4 | 0 | 4 | 0.3 | [0, 0, 0.1] | 6 |
| WL | ELWL | 2 | 5 | 0.3 | 0.8 | [0, 0, 0] | 2 |
| NL | NLSL | 2 | 2 | 1.3 | 0.6 | [0, 0, 0] | 3 |
| SL | NLSL | 2 | 10 | -1.3 | 0.2 | [0, 0, 0] | 4 |

### Phase Aggregation

Phase ranking by aggregated signal pressure: ELWL > NTST > NLSL > ETWT
Highest pressure phase from computed indicators: ELWL

| Phase | Movement | Early Queue Length | Downstream Queue Length | Pressure | Avg Waiting Time(minutes) | Segment Veh Count | Vehicle Count |
| --- | --- | --- | --- | --- | --- | --- | --- |
| ETWT | ET,WT | 0 | 12 | -4 | 0 | [0, 3, 2] | 5 |
| NTST | NT,ST | 2 | 5 | 0.3 | 0 | [1, 4, 12] | 19 |
| ELWL | EL,WL | 6 | 5 | 4.3 | 0.4 | [0, 0, 2] | 8 |
| NLSL | NL,SL | 4 | 12 | 0 | 0.4 | [0, 0, 3] | 7 |

## Adjacent Intersection Context

| Direction | Intersection ID | ET | WT | NT | ST | EL | WL | NL | SL | Related Movements |
| --- | --- | --- | --- | --- | --- | --- | --- | --- | --- | --- |
| East | intersection_1_3 | 11, 0.5 | 0, 0.1 | 4, 0.1 | 14, 0.2 | 10, 0.2 | 2, 0 | 4, 0.1 | 3, 0.1 | ET,SL -> ET,EL |
| West | intersection_3_3 | 6, 0.2 | 1, 0 | 0, 0.1 | 0, 0.1 | 4, 0.1 | 2, 0 | 0, 0 | 3, 0.1 | WT,NL -> WT,WL |
| North | intersection_2_2 | 5, 0.2 | 5, 0.2 | 0, 0.1 | 0, 0.1 | 1, 0 | 2, 0.1 | 2, 0 | 5, 0.1 | NT,EL -> NT,NL |

## Recent History

Recent phases: NLSL -> ETWT -> NTST -> ETWT
Last phase: ETWT

| Phase | Queue Change | Pressure Change | Waiting Change | Service Gap | Trend Summary |
| --- | --- | --- | --- | --- | --- |
| ETWT | 6 -> 0 -> 7 -> 0 | 5.7 -> -2.3 -> 4.3 -> -4 | 0.2 -> 0 -> 0.1 -> 0 min | 0 steps | recently served and clearly relieved |
| NTST | 7 -> 11 -> 0 -> 2 | 5.3 -> 9.3 -> -2.7 -> 0.3 | 0.3 -> 0.4 -> 0 -> 0 min | 1 steps | recently served and clearly relieved |
| ELWL | 1 -> 2 -> 3 -> 6 | -0.7 -> 0.3 -> 0.3 -> 4.3 | 0.1 -> 0.2 -> 0.3 -> 0.4 min | 5 steps | unserved for long and demand is worsening |
| NLSL | 0 -> 2 -> 4 -> 4 | -0.3 -> -0.3 -> 1.3 -> 0 | 0 -> 0.1 -> 0.2 -> 0.4 min | 3 steps | demand is increasing |

## Output Prompt
Reason step-by-step via seven fixed Level-2 heading sections:
Phase Mapping, Local Demand, Downstream, Neighbor Coordination, Flow Trend, Fairness, and Final Answer.

Each reasoning section should contain no more than three atomic <claim> tags.
Only mathematically derivable conclusions following the unified syntax may be placed within <claim>.
The final selected phase must be enclosed by <phase></phase>.
     \end{lstlisting}
    \caption{ The prompt template  }
    \label{tab:r0_template}
\end{table*}
\twocolumn

\onecolumn
\begin{table*}[t]
    \centering
    \small
    \begin{lstlisting}
## Phase Mapping
Phase mapping links each valid signal phase to its corresponding traffic movements. The phases are ranked by aggregated pressure to identify priority.
<claim>RANK[Pressure][Phase]=ELWL > NTST > NLSL > ETWT</claim>
<claim>MAX[Early Queue Length][Phase]=ELWL</claim>
<claim>MAX[Avg Waiting Time][Phase]=ELWL</claim>

## Local Demand
ELWL exhibits the highest local demand with an aggregated pressure of 4.3 and an early queue length of 6, both exceeding all other phases. WL's average waiting time of 0.8 minutes is the highest among movements.
<claim>[ELWL][Pressure]=4.3</claim>
<claim>[ELWL][Early Queue Length]=6</claim>
<claim>[WL][Avg Waiting Time]=0.8</claim>

## Downstream
Several downstream segments show high queue lengths that could limit discharge. SL and ET each have a downstream queue of 10, while WL has a downstream queue of 5, indicating potential congestion but still within manageable limits for ELWL's left-turn movement.
<claim>[SL][Downstream Queue Length]=10</claim>
<claim>[ET][Downstream Queue Length]=10</claim>
<claim>[WL][Downstream Queue Length]=5</claim>

## Neighbor Coordination
Adjacent intersections provide context for inflow and outflow. West-bound WL has a low queue of 2, allowing discharge. East-bound ET shows a high occupancy of 0.5, suggesting some downstream constraint, and North-bound NT has low occupancy of 0.1.
<claim>ADJ[West][WL][Queue]=2</claim>
<claim>ADJ[East][ET][Occupancy]=0.5</claim>
<claim>ADJ[North][NT][Occupancy]=0.1</claim>

## Flow Trend
ELWL has been unserved for 5 steps with worsening demand and increasing pressure and queue, indicating urgent need for service.
<claim>TREND[ELWL][Service Gap]=5 steps (unserved)</claim>
<claim>TREND[ELWL][Pressure Change]=increasing</claim>
<claim>TREND[ELWL][Queue Change]=increasing</claim>

## Fairness
WL holds the maximum average waiting time among all movements, but no movement exceeds the 10-minute threshold. The service gap ranking confirms ELWL has been waiting longest.
<claim>MAX[Avg Waiting Time][Movement]=WL</claim>
<claim>No [Movement] has [Avg Waiting Time]>=10</claim>
<claim>RANK[Service Gap][Phase]=ELWL > NLSL > NTST > ETWT</claim>

## Final Answer
<phase>ELWL</phase>
     \end{lstlisting}
    \caption{ Example of ProcessLight response.  }
    \label{tab:r0_response}
\end{table*}
\twocolumn

\end{document}